\documentclass{article} %
\usepackage{setup/iclr2024_conference,times}

\usepackage[compress,numbers]{natbib}
\usepackage[linkcolor=brown,colorlinks=True,citecolor=carmine]{hyperref}
\usepackage{url}
\definecolor{citecolor}{HTML}{0071bc}
\hypersetup{
    colorlinks,
    linkcolor={citecolor},
    citecolor={citecolor},
    urlcolor={citecolor}
}
\usepackage{booktabs}
\usepackage{multirow}
\usepackage{graphicx}
\usepackage{tabularx}
\usepackage{algorithm}
\usepackage{algpseudocode}
\usepackage{textcomp}
\usepackage{wrapfig}
\usepackage{subcaption}
\usepackage[labelfont=bf,font={small}]{caption}
\usepackage{authblk}
\usepackage{enumitem}

\title{Zero-Shot Robotic Manipulation with\\ Pretrained Image-Editing Diffusion Models}

\makeatletter
\renewcommand\AB@affilsepx{\quad \protect\Affilfont}
\makeatother

\author[1]{Kevin Black$^*$}
\author[1]{Mitsuhiko Nakamoto$^*$}
\author[1]{Pranav Atreya}
\author[1]{Homer Walke}
\author[2,3]{\authorcr Chelsea Finn}
\author[1,3]{Aviral Kumar}
\author[1,3]{Sergey Levine}
\affil[1]{University of California, Berkeley}
\affil[2]{Stanford University}
\affil[3]{Google DeepMind \protect\\}
\affil[*]{Equal contribution}

\usepackage{amsmath,amsfonts,bm}

\def\eqref#1{equation~\ref{#1}}

\def\1{\bm{1}}

\DeclareMathAlphabet{\mathsfit}{\encodingdefault}{\sfdefault}{m}{sl}
\SetMathAlphabet{\mathsfit}{bold}{\encodingdefault}{\sfdefault}{bx}{n}

\newcommand{\bs}{\mathbf{s}}
\newcommand{\ba}{\mathbf{a}}

\newcommand{\methodname}{SuSIE}

\newcommand{\figtitle}[1]{\textbf{(#1)}}

\iclrfinalcopy

\usepackage{color}

\begin{document}

\maketitle

\begin{abstract}
   
If generalist robots are to operate in truly unstructured environments, they need to be able to recognize and reason about novel objects and scenarios. Such objects and scenarios might not be present in the robot's own training data. We propose \methodname, a method that leverages an image-editing diffusion model to act as a high-level planner by proposing intermediate subgoals that a low-level controller can accomplish. Specifically, we finetune InstructPix2Pix on video data, consisting of both human videos and robot rollouts, such that it outputs hypothetical future ``subgoal'' observations given the robot's current observation and a language command. We also use the robot data to train a low-level goal-conditioned policy to act as the aforementioned low-level controller. We find that the high-level subgoal predictions can utilize Internet-scale pretraining and visual understanding to guide the low-level goal-conditioned policy, achieving significantly better generalization and precision than conventional language-conditioned policies.
We achieve state-of-the-art results on the CALVIN benchmark, and also demonstrate robust generalization on real-world manipulation tasks, beating strong baselines that have access to privileged information or that utilize orders of magnitude more compute and training data.
The project website can be found at \url{http://rail-berkeley.github.io/susie}.
\end{abstract}

\vspace{-0.2cm}
\section{Introduction}
\vspace{-0.2cm}

A useful generalist robot must be able to --- much like a person --- recognize and reason about novel objects and scenarios it has never encountered before. For example, if a user instructs the robot to ``hand me that jumbo orange crayon,'' it ought to be able to do so even if it has never interacted with a jumbo orange crayon before. In other words, the robot needs to possess not only the physical capability to manipulate an object of that shape and size but also the semantic understanding to reason about an object outside of its training distribution. As much as robotic manipulation datasets have grown in recent years, it is unlikely that they will ever include every conceivable instance of objects and settings, any more so than the life experiences of a person ever include physical interactions with every type of object. While these datasets contain more than enough examples of manipulating elongated cylindrical objects, they lack the broad semantic knowledge necessary to ground the \emph{particular} objects that robots will encounter during everyday operation.

How can we imbue this semantic knowledge into language-guided robotic control? One approach would be to utilize models pretrained on vision and language to initialize different components of the robotic learning pipeline. Recent efforts, for example, initialize robotic policies with pretrained vision-language encoders~\citep{rt2} or apply pretrained models to semantic scene augmentation~\citep{genaug,rosie}. While these methods bring semantic knowledge into robot learning, it remains unclear if these approaches realize the full potential of Internet pretraining in improving robotic policy execution at every level, or whether they simply improve the high-level visual generalization of the policy.

\begin{figure}
    \centering
    \vspace{-0.2cm}
    \includegraphics[width=0.87\linewidth]{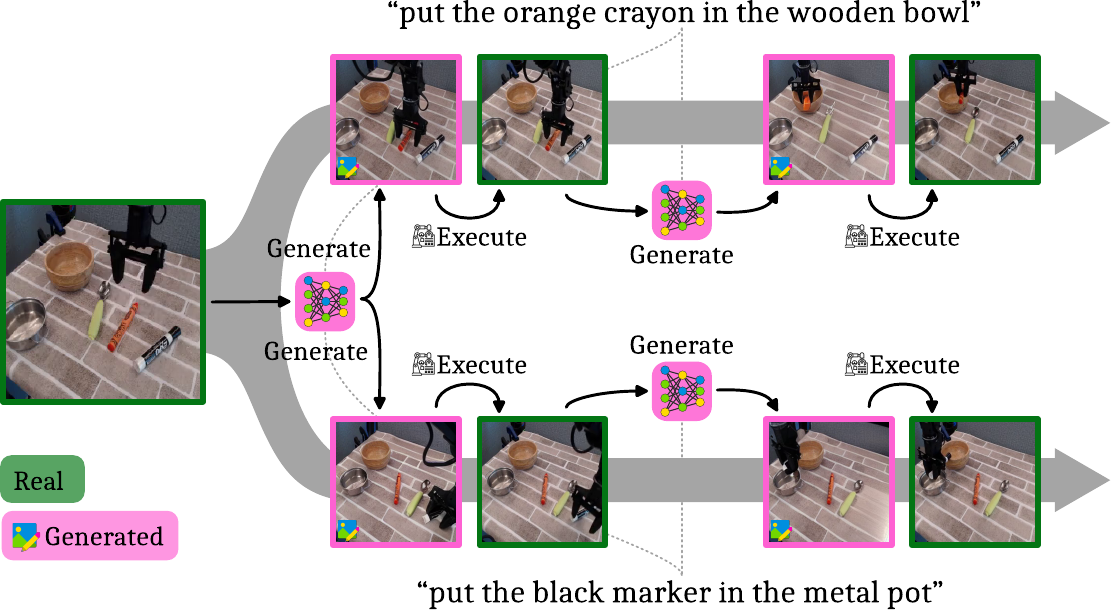}
    \caption{\figtitle{\methodname} Our method leverages a pretrained image-editing model to generate future subgoals based on language commands. A low-level goal-reaching policy then executes the actions needed to reach each subgoal. Alternating this loop enables us to solve the task.}
    \vspace{-0.4cm}
    \label{fig:teaser}
\end{figure}

We propose an approach for leveraging pretrained image-editing models to enable generalizable robotic manipulation. We finetune the image-editing model on video data such that, given the current frame and a language description of the current task, the model generates a hypothetical \textit{future} frame.
This does not require the model to precisely understand the intricacies of the robot's low-level dynamics and therefore should facilitate transfer from other data sources (e.g., human videos) where the low-level physical interactions and precise embodiment do not match. At test time, we employ a low-level goal-reaching policy trained on robot data to reach this hypothetical future frame; this policy, in turn, only needs to infer visuo-motor relationships to determine the correct actuation and does not need to understand the overlying semantics. Furthermore, such subgoals can simplify the task by inferring likely poses for the arm at intermediate substeps, such as the pose corresponding to grasping an object (see Figure~\ref{fig:teaser}). In fact, we observe in our experiments that even when existing approaches possess sufficient semantic understanding to solve a task, they often fail due to imprecise localization of obstacles and objects; following the generated subgoals enables our method to perform well in such scenarios. Much like a person first constructs a high-level plan to complete a task before deferring to muscle memory for the low-level control, our method can also be viewed as first running a high-level planner with integrated semantic reasoning \textit{and} visual understanding before deferring to a low-level controller to execute the plan.

The main contribution of our work is \textbf{SU}bgoal \textbf{S}ynthesis via \textbf{I}mage \textbf{E}diting (\textbf{\methodname}), a simple and scalable method for incorporating semantic information from pretrained models to improve robotic control. A pretrained image-editing model is used with minimal modification, requiring only finetuning on video data. The low-level goal-conditioned policy is trained with standard supervised learning, and faces the comparatively easy problem of reaching nearby image subgoals; this typically only requires attending to a single object or the arm position, ignoring most parts of the scene.
Together, this approach achieves state-of-the-art results on the CALVIN benchmark and solves real robot control tasks involving novel objects, novel distractors, and even novel scenes. It outperforms all prior approaches on these real-world tasks, including oracle baselines with privileged information and RT-2-X~\citep{rt2x} --- a 55 billion parameter model trained on Internet-scale vision-language data as well as an order of magnitude more robot data than \methodname.

\vspace{-0.2cm}
\section{Related Work}
\vspace{-0.2cm}
\label{sec:related}
\textbf{Incorporating semantic information from pretrained vision-language models.} Prior works that incorporate semantic information from pretrained vision-language models (VLMs) into robot learning can be classified into two categories. The first category aims to improve visual scene understanding in robot policies with information from VLMs. For instance, GenAug~\citep{genaug}, ROSIE~\citep{rosie}, DALL-E-Bot~\citep{dall-e-bot}, and CACTI~\citep{cacti} use text-to-image generative models to produce semantic augmentations of a given scene with novel objects and arrangements and train the robot policy on the augmented data to enable it to perform well in a similar scene. MOO~\citep{moo} uses a pretrained object detector to extract bounding boxes to guide the policy towards the target object. Other works directly train language and image-based policies~\citep{brohan2022rt,rt2,cliport}, some using VLM initializations~\citep{palme,clip}, to produce action sequences.    

While these approaches do utilize pretrained models, we find in our experiments that pretraining using VLMs~\citep{rt2} does not necessarily enhance low-level policy execution in the sense that learned policies often localize the object or move the gripper imprecisely (see Figure~\ref{fig:rollouts}). On the other hand, our approach is able to incorporate benefits of pretraining into low-level control by synthesizing subgoals that carefully steer the motion of the low goal reaching policy, improving its precision.
While our approach can be directly applied in \emph{unstructured} open-world settings, applying methods from \citep{genaug,rosie,cacti} requires additional prior information, such as object bounding boxes or 3D meshes. This significantly restricts their applicability to scenarios where this additional information is not available: for example, GenAug~\citep{genaug} is not applicable in our experiments since 3D object meshes for new target objects were not available.
Distinct from our approach for utilizing generative models, other works design representation learning objectives for vision-language pretraining for control~\citep{r3m,liv,voltron,vptr}, but these methods still need to utilize limited amounts of data from the target task to learn a policy. In contrast, our approach solves the target task in zero-shot, without any task-specific data.

The second category of approaches incorporates semantic information from pretrained models for informing low-level controllers. Typically, these prior methods use pretrained models to imagine visual~\citep{unipi,hip} or textual plans~\citep{saycan,huang2022language,huang2022inner,liang2023code},
which then inform a low-level robot control policy. Of these, methods that train a low-level policies conditioned on text suffer from a grounding problem, which our approach circumvents entirely since the low-level policy only sees images. 

Perhaps the most related methods to our approach in this second category are UniPi~\citep{unipi} and HiP~\citep{hip}, which train video models to generate a sequence of frames achieving the target task, and then extract robot actions by training an inverse dynamics model. Our approach does \emph{not} attempt to generate full videos (i.e., \emph{all} frames in a rollout), but only the next waypoint that a low-level policy must achieve to solve the commanded task. While this difference might appear small, it has major implications: modeling an entire video puts a very high burden on the generative model, requiring the frames to obey strict physical consistency. Unfortunately, we find that open-source video models often produce temporally inconsistent frames, which confuse the low-level controller (see Appendix~\ref{appendix:unipi_examples}). 
Our method provides more freedom to the low-level controller to handle the physical aspects of the task over a longer time interval while guiding it at a level that is suitable to the diffusion model's ability to preserve physical plausibility. 
In our experiments, we find that our method significantly improves over an open-source reimplementation of UniPi~\citep{unipi}.

\textbf{Classical model-based RL and planning with no pretraining.} The idea behind our approach is also related to several methods in the deep RL literature that do not use pretrained models and generally do not study language-guided control. For instance, \citep{hafner2019dream,lee2020stochastic,yu2021combo,wu2023daydreamer,Rafailov2020LOMPO,hafner2023mastering} train action-conditioned dynamics models and run RL in the model. While our approach also models multi-step dynamics, our model is not conditioned on an action input. Removing the dependency on an action input enables us to de-couple the finetuning of the (large) image-editing model from the policy entirely, improving simplicity and time efficiency. APV~\citep{seo2022reinforcement} trains an action agnostic dynamics model from videos but finetunes it in a loop with the policy, and hence, does not enjoy the above benefits.   
Finally, these model-based RL methods do not exhibit zero-shot generalization abilities to new tasks, which is an important capability that our method enjoys. Our approach is also related to several video prediction methods~\citep{ebert2018visual,lee2018stochastic,babaeizadeh2021fitvid,villegas2019high} but utilizes a better neural network architecture (i.e., diffusion models instead of LSTMs and CNNs). Most related is to our method is hierarchical visual foresight (HVF)~\citep{nair2019hierarchical}: while HVF uses MPC to find an action, our approach uses a goal-reaching policy thereby eliminating the cost of running MPC with large dynamics models. 

Our approach is also related to prior works that use generative models for planning in a single-task setting, with no pretraining. Trajectory transformer (TT)~\citep{janner2021sequence}, decision transformer (DT)~\citep{chen2021decision}, and their extensions condition the policy on the target return or goal. While diffusion-based variants of these methods~\citep{janner2022diffuser,ajay2023is} use diffusion models like our approach, they still require training data from the target task to learn a policy, unlike our zero-shot planning approach.

\vspace{-0.25cm}
\section{Preliminaries and Problem Statement}
\label{sec:prelim}
\vspace{-0.25cm}

We consider the problem setting of language-conditioned robotic control. Specifically, we want a robot to accomplish a task described by a novel language command. We divide sources of data into 3 categories: language-labeled video clips $\mathcal{D}_l$ that do not include robot actions; language-labeled robot data $\mathcal{D}_{l,a}$ that include both language labels and robot actions; and unlabeled robot data $\mathcal{D}_{a}$ that include only actions (e.g., play data).

Formally, we define $\mathcal{D}_{l,a} = \{ \left(\tau^1, l^1 \right), \left( \tau^2, l^2 \right), \cdots, \left( \tau^N, l^N \right) \}$, where each trajectory $\tau^n$ consists of a sequence of images (or states) $\bs^n_i \in \mathcal{S}$ and actions $\ba^n_i \in \mathcal{A}$ that were executed while collecting this data,
i.e. $\tau^n = \left( \bs^n_0, \ba^n_0, \bs^n_1, \ba^n_2, \cdots \right)$, following the standard assumptions of a Markov decision \mbox{process}. $l^n$ is a natural language command describing the task accomplished in the trajectory. $\mathcal{D}_{l}$ and $\mathcal D_a$ are organized similarly, but are missing either actions $\ba^n_i$ or language annotations $l^n$, respectively. At test time, given a new scene $\bs_0^\text{test}$ and a new natural language description $l^\text{test}$ of a task, we evaluate a method in terms of its success rate at accomplishing this task starting from $\bs_0^\text{test}$.

\vspace{-0.1cm}
\section{\methodname: Subgoal Synthesis via Image Editing}
\vspace{-0.1cm}

Our goal is to utilize semantic information from the Internet to improve language-guided robot control in the presence of novel environments, scenes, and objects. How can we do this when models trained on general-purpose Internet data do not provide guidance in selecting low-level actions? Our key insight is that we can effectively leverage the capabilities of the pretrained model if we decouple the robot control problem into two phases: \textbf{\textsc{(i)}} generating subgoals that would need to be attained to succeed at the task, and \textbf{\textsc{(ii)}} learning low-level control policies for reaching these generated subgoals. Our method incorporates semantic information from both text-image pretraining on Internet data as well as non-robot video data in Phase (\textsc{i}) by finetuning a text-guided image-editing model on $\mathcal D_l \cup \mathcal D_{l,a}$. Phase (\textsc{ii}) is accomplished via a goal-conditioned policy trained on the robot data $\mathcal D_{l,a} \cup \mathcal D_a$. We describe each of these phases below and then summarize the resulting algorithm.

\vspace{-0.2cm}
\subsection{Phase (\textsc{i}): Synthesizing Subgoals From Image-Editing Models}
\vspace{-0.1cm}

The primary component of our method is a generative model that, given a target task specified by natural language, can guide a low-level controller to a state that advances the task. One way to accomplish this is to train a generative model to produce an immediate next subgoal image. We can then incorporate semantic information from the Internet into our algorithm by initializing this generative model with a suitable pretrained initialization, followed by finetuning it on multi-task, diverse video data consisting of robot rollouts as well as other videos from the Internet.

\emph{What is a good pretrained initialization for this model?} Our intuition is that, since accomplishing a task is equivalent to ``editing'' the pixels of an image of the robot workspace under constraints prescribed by the language command, a good pretrained initialization may be provided by a text-guided image-editing model. We instantiate our approach with Instruct Pix2Pix~\citep{instructpix2pix}, though other image-editing models could also be used. Formally, this model is given by $p_\theta(\bs_\text{edited} \mid \bs_\text{orig}, l)$. Then, using the dataset $\mathcal D_l \cup \mathcal D_{l,a}$ of language-labeled video clips and robot trajectories, we finetune $p_\theta$ to produce valid subgoals $\bs_\text{edited}$ given an initial image $\bs_\text{orig}$ and a language label $l$.
Formally, the training objective is given by
\begin{align}
\label{eqn:finetuning_phase1}
    \max_\theta~~ \mathbb{E}_{~(\tau^n, l^n) \sim \mathcal D_l \cup \mathcal D_{l,a};~~ \bs_i^n \sim \tau^n;~~ j \sim q(j \mid i)} \left[ \log p_\theta \left(\bs_j^n  \mid \bs_i^n, l^n\right) \right],
\end{align} 
where $q(j \mid i)$ is a distribution of our choosing that controls what subgoals the model is trained to produce. We want the diffusion model to generate subgoals that are a consistent distance in the future from the current state --- close enough to be reachable by the low-level policy, but far enough to achieve significant progress on the task. As such, we choose dataset-dependent hyperparameters $k_\text{min}$ and $k_\text{max}$, and set $q$ to select subgoals uniformly between $k_\text{min}$ and $k_\text{max}$ steps in the future, i.e.,: 
\begin{align*}
    q(j \mid i) = U\big(j; \; [i + k_{\text{min}}, i + k_{\text{max}})\big).
\end{align*}

\vspace{-0.2cm}
\subsection{Phase (\textsc{ii}): Reaching Generated Subgoals with Goal-Conditioned Policies}
\label{method:GCBC}
\vspace{-0.2cm}
In order to utilize the finetuned image-editing model to control the robot, we additionally need to train a low-level controller to select suitable robot actions. In this section, we present the design of our low-level controller, followed by a full description of our test-time control procedure. Since the image-editing model in \methodname\ produces images of future subgoals conditioned on natural language task descriptions, our low-level controller can simply be a language-agnostic goal-reaching policy.

\textbf{Training a goal-reaching policy.} Our goal-reaching policy is parameterized as $\pi_\phi(\ba \mid \bs_i, \bs_j)$, where $\bs_j$ is a future frame that the policy intends to reach by acting at $\bs_i$. At test time, we only need the low-level policy to be proficient at reaching close-by states that lie within $k_\text{max}$ steps of the current state, since the image-editing model from Phase (\textsc{i}) is trained to produce subgoals within $k_\text{max}$ steps of any state. To train this policy, we run goal-conditioned behavioral cloning (GCBC) on the robot data $\mathcal D_{l,a} \cup \mathcal D_{a}$. Formally, the training objective is given by
\begin{align}
    \label{eqn:gcbc_training}
    \max_\phi~~ \mathbb{E}_{~\tau^n \sim \mathcal{D}_{l,a} \cup \mathcal{D}_a; ~(\bs_i^n, \ba_i^n) \sim \tau^n; ~ j \sim U([0, k_\text{max} + k_\delta))} \left[ \log \pi_\phi(\ba_i^n  \mid  \bs_i^n, \bs_j^n) \right],
\end{align}
where $k_\delta$ is another hyperparameter that provides a small amount of overhead, since the image-editing model is not perfect and may not always produce subgoals that are reachable in less than $k_\text{max}$ steps, especially for unseen tasks.

\textbf{Test-time control with $\pi_\phi$ and $p_\theta$.} Once both the goal-reaching policy $\pi_\phi$ and the subgoal generation model $p_\theta$ are trained, we utilize them together to solve new manipulation tasks based on user-specified natural language commands. Given a new scene $\bs_0^\text{test}$ and a language command $l^\text{test}$, \methodname\ attempts to solve the task by iteratively generating subgoals and commanding the low-level policy to reach these subgoals.
At the start, we sample the first subgoal $\widehat{\bs}_+ \sim p_\theta(\bs_+ \mid \bs_0^\text{test}, l^\text{test})$. Once the subgoal is generated, we then roll out the goal-reaching policy $\pi_\phi$ conditioned on $\widehat{\bs}_{+}$ for $k_\text{test}$ timesteps where $k_\text{test}$ is a test-time hyperparameter. After $k_\text{test}$ timesteps, we refresh the subgoal by sampling from the subgoal generation model again and repeat the process. In practice, for computational efficiency, we set $k_\text{test}$ to be similar to the corresponding $k_\text{max}$ used with the robot data and found this to be sufficient for obtaining good performance. However, given an unlimited computational budget, conventional wisdom would suggest that regenerating subgoals more often would lead to more robust control. Pseudocode for test-time control is provided in Algorithm~\ref{algo:planning}.          

\begin{algorithm}
\caption{\methodname: Zero-Shot, Test-Time Execution}
\label{algo:planning}
\begin{algorithmic}[1]
\Require{Subgoal model $p_\theta(\bs_{+} \mid  \bs_t, l)$, policy $\pi_\phi \left(\ba \mid \bs_{t}, \bs_{+} \right)$, time limit $T$, subgoal sampling interval $k_\text{test}$, initial state $\bs_0^\text{test}$}, language command $l^\text{test}$
\State $t \gets 0$
\While{$t \leq T$}
    \State Sample $\widehat \bs_+ \sim p_\theta(\bs_{+}  \mid  \bs^\text{test}_t, l^\text{test})$ \Comment{Generate a new subgoal every $k_\text{test}$ steps}
    \For{$j = 1$ to $k_\text{test}$}
        \State Sample $\ba_t \sim \pi_\phi  \left( \ba \mid \bs^\text{test}_t, \widehat \bs_+ \right)$ \Comment{Predict the action from current state and subgoal}
        \State Execute $\ba_t$
        \State $\bs^\text{test}_{t+1} \gets \text{robot observation}$
        \State $t \gets t + 1$
    \EndFor
\EndWhile
\end{algorithmic}
\end{algorithm}

\vspace{-0.2cm}
\subsection{Implementation Details}
\vspace{-0.2cm}

In Phase (\textsc{i}), we utilize the pretrained initialization from the InstructPix2Pix model~\citep{instructpix2pix}. We implement Equation~\ref{eqn:finetuning_phase1} using the standard variational lower bound objective for training diffusion models \citep{ho2020denoising}. Our diffusion model and policy operate on images of size \(256 \times 256\). To ensure that this model pays attention to the input image and the language command, we apply classifier-free guidance~\citep{classifier-free} separately to both the language and the image, following InstructPix2Pix. 
To obtain a robust goal-reaching policy in Phase (\textsc{ii}), we use a diffusion policy~\citep{chi2023diffusion,hansen2023idql} that predicts chunks of 4 actions and performs temporal averaging over these predictions at test time~\citep{zhao2023learning}. More details about the training hyperparameters and architecture are provided in Appendix \ref{appendix:ours}.

\vspace{-0.2cm}
\section{Experimental Evaluation}
\vspace{-0.2cm}

The goal of our experiments is to evaluate the efficacy of \methodname\ at improving generalization and the low-level policy execution in open-world robotic manipulation tasks. To this end, our experiments aim to study the following questions:
\begin{enumerate}
    \item Can \methodname\ solve a task in a novel environment, with novel objects, and with a novel language command, in zero-shot?
    \item Does \methodname\ exhibit an elevated level of precision and dexterity compared to other approaches that do not use subgoals?
    \item How crucial is pretraining on Internet data, as well as cotraining on non-robot video data, for attaining zero-shot generalization?
\end{enumerate}

To answer these questions, our experiments compare \methodname\ to several prior methods including state-of-the-art approaches for training language-conditioned policies that leverage pretrained vision-language models in a variety of ways. 

\vspace{-0.2cm}
\subsection{Experimental Scenarios and Comparisons}
\vspace{-0.2cm}

\begin{figure}[b!]
    \centering
    \includegraphics*[width=1.0\textwidth]{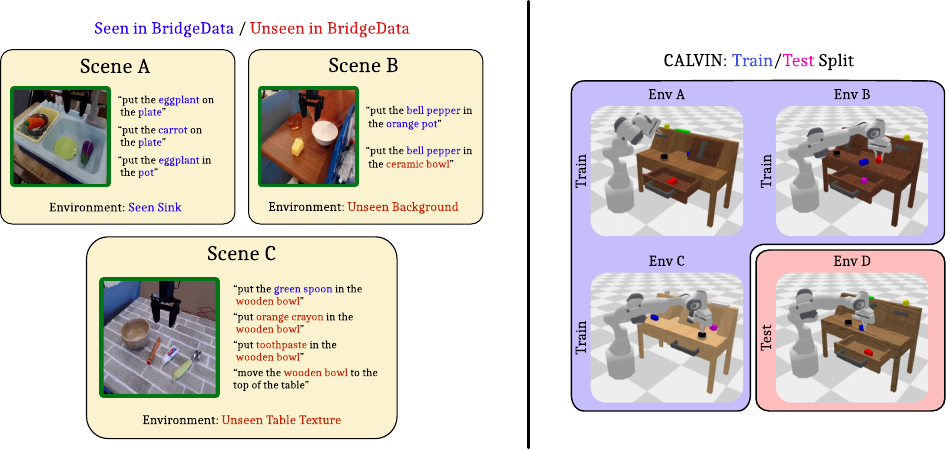}
    \caption{\figtitle{Experimental setup}
    \figtitle{L} We evaluate our method on 9 tasks across 3 real-world scenes. In terms of semantic generalization, the scenes become progressively more difficult, due to both an increasing visual departure from the robot training data and an increasingly confounding mixture of both seen and unseen objects.
    \figtitle{R} In simulation, we evaluate our method in the zero-shot setting of the CALVIN benchmark, which involves training on 3 environments (A, B, and C) and testing on a 4th environment (D). The environments differ in table texture, positioning of furniture elements, and possible configurations of various colored blocks. Each environment comes with 34 language-specified tasks.
    }
    \label{fig:scenes}
    \vspace{-0.4cm}
\end{figure}

\textbf{Real-world experimental setup and datasets.} We conduct our real-robot experiments on a WidowX250 robot platform. Our dataset is BridgeData V2~\citep{bridgev2}, a large and diverse dataset of robotic manipulation behaviors designed for evaluating open-vocabulary instructions. The dataset contains over 60k trajectories, 45k of which are language-labeled, which we use as our language-labeled robot dataset $\mathcal D_{l,a}$. We use the remaining 15k trajectories as our action-only dataset $\mathcal D_a$.

Our video-only dataset $\mathcal D_l$ is the Something-Something dataset \citep{goyal2017something}, a dataset consisting of short video clips of humans manipulating various objects. We chose Something-Something because it primarily contains examples of object manipulation with a still camera frame and hence exhibits a smaller domain gap to the robot data collected with an over-the-shoulder camera compared to other video datasets that contain substantial egocentric motion. We filter the Something-Something dataset using its labels to remove trajectories that contain egocentric motion or otherwise lack significant manipulation behavior, producing a final dataset size of approximately 75k video clips.

Our evaluations present three different scenes~(Figure~\ref{fig:scenes}) designed specifically to test the ability of various methods at different levels of open-world generalization: \textbf{Scene A:} this scene includes an environment and objects that are well-represented in BridgeData V2; \textbf{Scene B:} this scene is situated in an environment with a seen tabletop but a novel background and distractors, where the robot must move a seen object (bell pepper) into a choice of seen container (orange pot) or unseen container (ceramic bowl); and \textbf{Scene C:} this scene includes a table texture unlike anything in BridgeData V2 and requires manipulating both seen and unseen objects. Semantically, Scene C is the most challenging since the robot needs to carefully ground the language command to identify the correct object while resisting its affinity for an object that is well-represented in the robot data (the spoon). Scene B also requires semantic grounding to distinguish between the seen and unseen receptacles, while adding the additional challenge of manipulating the plastic bell pepper --- this object requires particular precision to grasp due to being light, smooth, and almost as wide as the gripper.

\textbf{Simulation tasks.} We run our simulation experiments in CALVIN~\citep{calvin}, a benchmark for long horizon, language-conditioned manipulation. CALVIN consists of four simulated environments, labeled A through D, and each environment comes with a dataset of human-collected play trajectories. Approximately $35\%$ of the play data is annotated with language, which we use as our language-labeled robot dataset $\mathcal D_{l,a}$. We use the remaining $65\%$ of the play data as our action-only robot dataset $\mathcal D_a$, and do not use any video-only dataset $\mathcal D_l$ in our simulation experiments.

Each environment consists of a Franka Emika Panda robot arm positioned next to a desk with various manipulatable objects, including a drawer, sliding cabinet, light switch, and various colored blocks. Environments are differentiated by their table textures, positions of the furniture objects, and configurations of colored blocks. With this benchmark, we study the most challenging zero-shot multi-environment scenario: training on A, B, and C, and testing on D. We follow the evaluation protocol from \citet{calvin}. During evaluation, a policy is given 360 timesteps to complete a chain of five language instructions. We provide more details about our experimental setup in Appendix~\ref{appendix:experimental_setup}.

\textbf{Comparisons.} Our experiments cover methods that utilize pretrained models of vision and language in language-guided robot control in a variety of ways. Several prior methods tackle language-based robotic control as discussed in Section~\ref{sec:related}. In our experiments, we choose to compare to a representative subset of these prior methods to maximally cover the possible set of comparisons. In our real-world experiments, we compare to the following methods:
\begin{itemize}[leftmargin=2em]
\item \textbf{RT-2-X}~\citep{rt2x}, a 55 billion parameter vision-language model finetuned on the full Open X-Embodiment dataset to produce robot actions. The robot training data includes all of BridgeData V2, as well as a vast quantity of additional robot manipulation data totaling over 1.1 million trajectories.

\item \textbf{MOO}~\citep{moo}, which utilizes pretrained object detectors to obtain 2D localization information for the manipulation targets and trains a language-conditioned behavioral cloning (LCBC) policy with this extra information. We re-implement MOO by using a large language model to extract the manipulation targets from the unstructured language labels in BridgeData V2, then OWL-ViT~\citep{owl-vit} to extract 2D bounding boxes for these objects. Unlike MOO, at test time we manually give the policy the ground truth 2D bounding boxes of the manipulation targets.

\item \textbf{UniPi}~\citep{unipi}, which finetunes a language-conditioned video prediction model on robot data. Since the original UniPi model utilized proprietary pretrained initializations that are not available publicly, we replicated this method in two different ways: (1) using the UniPi model from HiP~\citep{hip}, which is a PVDM~\citep{pvdm} latent video diffusion model, and (2) by implementing our own image-space video diffusion model with a factorized spatial-temporal 3D UNet similar to \citet{vdm}. We found that (1) was not able to produce reasonable-looking videos in the real robot scene (Appendix~\ref{appendix:unipi_examples}), so we only evaluated (2) in our real-world experiments.

\item \textbf{LCBC}, which trains a low-level policy conditioned on the language labels in BridgeData V2. This method is broadly representative of prior language-conditioned behavioral cloning methods such as RT-1~\citep{brohan2022rt} and BC-Z~\citep{jang2022bc}. We use the same architecture and hyperparameters as the low-level policy in \methodname, as we found that they outperformed RT-1 on the BridgeData tasks.
\end{itemize}
We provide more details about the baselines and their implementations in Appendix~\ref{appendix:baselines}.

In our simulation experiments, we compare to additional methods previously studied on CALVIN. This includes methods that explicitly tackle long-horizon language-based control on CALVIN such as multi-context imitation (MCIL)~\citep{lynch2020language}, hierarchical universal language-conditioned policy (HULC)~\citep{hulc}, and improved variants of HULC~\citep{ge2023policy}. We also compare to other state-of-the-art methods from \citet{ge2023policy} that employ an identical training and evaluation protocol as our experiments, namely MdetrLC~\citep{kamath2021mdetr} and AugLC~\citep{pashevich2019learning}.

\vspace{-0.2cm}
\subsection{Can \methodname\ Perform Zero-Shot Robotic Manipulation?}
\vspace{-0.2cm}
\label{sec:zero_shot}

\textbf{Simulation results.} We present performance in Table~\ref{tab:calvin}, in terms of success rates (out of 1.0) for completing each language instruction in the chain. \methodname{} obtains superior zero-shot performance (train A, B, C $\to$ test D) compared to the previous state-of-the-art. We find that our reimplementation of UniPi achieves nontrivial success, but is still significantly outperformed by \methodname. 

\textbf{Real-world results.} We present performance of real-world evaluations in Table~\ref{tab:real}. \methodname\ achieves the best performance across the board, beating RT-2-X, a 55 billion parameter model trained on significantly more robot and Internet data. As expected, all methods perform well in Scene A, which is well-represented in the robot data. \methodname\ performs uniquely well in Scene B, as it is the only method that can consistently grasp the bell pepper.

In Scene C, while \methodname\ still achieves the best performance, RT-2-X comes in a close second. We hypothesize that this is because --- in contrast to Scene B --- all of the objects in Scene C are easy to grasp. Therefore, the low-level precision of the policy is less important, which is the primary weakness of RT-2-X, as we discuss further in the next section. Qualitatively, we observed that the failure cases in Scene C for both \methodname\ and RT-2-X were almost always imprecise manipulations (failed grasps or early dropping) rather than semantic misunderstanding; of the 4 objects, the toothpaste is the most difficult to grasp, which is why its success rate is the lowest. That is, both methods solve the semantic understanding component of the tasks, but \methodname's improved low-level precision allows it to perform slightly better.

The performance of LCBC is as expected: since it is trained only on BridgeData V2, it has no way of grounding novel objects, and often puts the bell pepper in the wrong receptacle (in Scene B) or attempts to grasp the wrong object (in Scene C). Surprisingly, it can recognize the wooden bowl in Scene C, even though that exact object does not appear in the training data; we hypothesize that the wooden bowl is distinctive enough to recognize from the word ``bowl'' alone combined with the many other bowls in BridgeData V2. MOO underperforms expectations, even with its privileged test-time information in the form of ground-truth bounding boxes. We hypothesize that, despite our best attempts, the target objects and bounding boxes extracted at train time are noisy due to the unstructured nature of BridgeData V2 and therefore only serve to confuse the policy (see Appendix~\ref{appendix:moo}). UniPi's poor performance is due to the video model often producing temporally inconsistent frames (see Appendix~\ref{appendix:unipi_examples}).

\begin{table}[t]
\centering
    {\begin{tabular}{l*{5}{>{\centering\arraybackslash}p{25pt}}}
        \toprule
        & \multicolumn{5}{c}{No. of Instructions Chained} \\
        \cmidrule{2-6}
        & 1 & 2 & 3 & 4 & 5 \\
        \midrule
        HULC~\citep{hulc} & 0.43 & 0.14 & 0.04 & 0.01 & 0.00 \\
        MCIL~\citep{lynch2020language} & 0.20 & 0.00 & 0.00 & 0.00 & 0.00 \\
        MdetrLC~\citep{ge2023policy} & 0.69 & 0.38 & 0.20 & 0.07 & 0.04 \\
        AugLC~\citep{ge2023policy} & 0.69 & 0.43 & 0.22 & 0.09 & 0.05 \\
        LCBC~\citep{bridgev2} & 0.67 & 0.31 & 0.17 & 0.10 & 0.06\\
        UniPi (HiP)~\citep{hip} & 0.08 & 0.04 & 0.00 & 0.00 & 0.00 \\
        UniPi (Ours)~\citep{unipi} & 0.56 & 0.16 & 0.08 & 0.08 & 0.04 \\
        \midrule
        \methodname~(Ours) & \textbf{0.87} & \textbf{0.69} & \textbf{0.49} & \textbf{0.38} & \textbf{0.26} \\
        \bottomrule
    \end{tabular}}
    \vspace{-0.2cm}
\caption{\figtitle{CALVIN benchmark performance} \methodname{} is able to chain together more instructions with a higher success rate than all prior methods in the zero-shot (train A, B, C $\to$ test D) setting. We provide an example rollout in Appendix~\ref{appendix:calvin}.}
\label{tab:calvin}
\vspace{-0.25cm}
\end{table}

\begin{table}
    \centering
    \begin{tabular}{llccccc}
        \toprule
        & Task & LCBC & MOO & UniPi & RT-2-X & \methodname\ (Ours) \\
        \midrule
        \multirow{4}{*}[-3pt]{Scene A\;\;} & Eggplant on plate & 0.9 & 0.4 & 0.0 & 0.3 & \textbf{1.0} \\
        & Carrot on plate & 0.4 & 0.3 & 0.0 & 0.6 & \textbf{0.9} \\
        & Eggplant in pot & 0.6 & \textbf{0.7} & 0.0 & 0.4 & \textbf{0.7} \\
        \cmidrule(lr){2-7}
        & Average & 0.63 & 0.47 & 0.0 & 0.43 & \textbf{0.87} \\
        \midrule
        \multirow{3}{*}[-3pt]{Scene B} & Bell pepper in pot & 0.1 & 0.0 & 0.0 & 0.0 & \textbf{0.5} \\
        & Bell pepper in bowl & 0.3 & 0.1 & 0.1 & 0.0 & \textbf{0.5} \\
        \cmidrule(lr){2-7}
        & Average & 0.20 & 0.05 & 0.05 & 0.00 & \textbf{0.50} \\
        \midrule
        \multirow{5}{*}[-3pt]{Scene C} & Toothpaste in bowl & 0.0 & 0.0 & 0.0 & 0.5 & \textbf{0.6} \\
        & Crayon in bowl & 0.0 & 0.0 & 0.0 & 0.9 & \textbf{1.0} \\
        & Spoon in bowl & 0.1 & 0.3 & 0.1 & 0.7 & \textbf{0.9}  \\
        & Bowl to top & 0.6 & 0.1 & 0.1 & 0.9 & \textbf{1.0}  \\
        \cmidrule(lr){2-7}
        & Average & 0.18 & 0.10 & 0.05 & 0.75 & \textbf{0.88} \\
        \bottomrule
    \end{tabular}
    \vspace{-0.25cm}
    \caption{\figtitle{Real-world performance} \methodname\ achieves the best success rate across the board, demonstrating both high precision as well as the ability to generalize to novel environments, objects, and language commands.}
    \label{tab:real}
    \vspace{-0.45cm}
\end{table}

\vspace{-0.3cm}
\subsection{Does \methodname{} Improve Precision and Low-Level Skill Execution?}
\label{sec:precision}

\begin{figure}[t]
    \centering
    \includegraphics[width=0.75\linewidth]{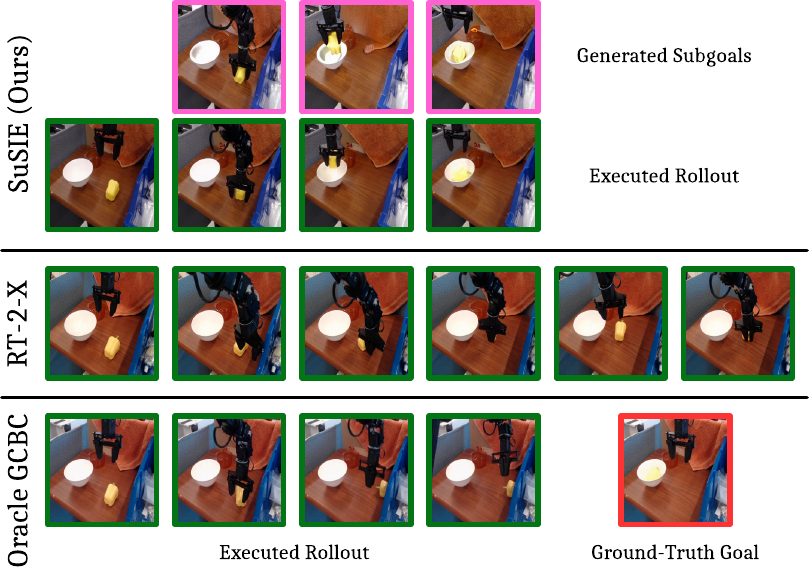}
    \caption{\figtitle{Visualizing rollouts} Visualized rollouts from \methodname, RT-2-X, and Oracle GCBC on the task ``put the yellow bell pepper in the ceramic bowl'' from Scene B. While RT-2-X and Oracle GCBC fail to grasp the object, the generated subgoals from \methodname\ precisely guide the low-level controller, improving low-level skill execution for difficult manipulation tasks.}
    \label{fig:rollouts}
\end{figure}

\begin{table}[tbp]
    \centering
    \begin{tabular}{llcc}
        \toprule
        & Task & Oracle GCBC & \methodname\ (Ours) \\
        \midrule
        \multirow{4}{*}[-3pt]{Scene A\;\;} & Eggplant on plate & 0.7 & \textbf{1.0} \\
        & Carrot on plate & 0.8 & \textbf{0.9} \\
        & Eggplant in pot & \textbf{1.0} & 0.7 \\
        \cmidrule(lr){2-4}
        & Average & 0.83 & \textbf{0.87} \\
        \midrule
        \multirow{3}{*}[-3pt]{Scene B} & Bell pepper in pot & 0.1 & \textbf{0.5} \\
        & Bell pepper in bowl & 0.0 & \textbf{0.5} \\
        \cmidrule(lr){2-4}
        & Average & 0.05 & \textbf{0.50} \\
        \midrule
        \multirow{5}{*}[-3pt]{Scene C} & Toothpaste in bowl & \textbf{0.7} & 0.6 \\
        & Crayon in bowl & \textbf{1.0} & \textbf{1.0} \\
        & Spoon in bowl & 0.8 & \textbf{0.9}  \\
        & Bowl to top & 0.6 & \textbf{1.0}  \\
        \cmidrule(lr){2-4}
        & Average & 0.78 & \textbf{0.88} \\
        \midrule
        \multirow{5}{*}[-3pt]{CALVIN} & Move slider right/left & 0.52 & \textbf{0.86} \\
        & Turn on/off light bulb & 0.74 & \textbf{0.96} \\
        & Open/close drawer & \textbf{1.00} & 0.98 \\
        & Turn on/off LED & 0.36 & \textbf{1.00} \\
        \cmidrule(lr){2-4}
        & Average & 0.66 & \textbf{0.95} \\
        \bottomrule
    \end{tabular}
    \vspace{-0.25cm}
    \caption{\figtitle{Comparison to GCBC with oracle goals} Executing generated subgoals improves performance even compared to executing privileged ground-truth final goals.}
    \label{tab:compare_to_oracle}
    \vspace{-0.45cm}
\end{table}

Our real-world and simulated results clearly demonstrate the efficacy of \methodname\ in executing novel language commands in a variety of scenes. In Section~\ref{sec:zero_shot}, we hypothesize that the advantage of \methodname\ is twofold: semantic generalization due to Internet and video pretraining, and low-level precision due to subgoal guidance. In this section, we aim to validate the latter half of this hypothesis. To this end, we train an \textbf{Oracle GCBC} policy, which is identical to the low-level policy in \methodname\ except that it is trained with a goal horizon that extends to the end of each trajectory. At test time, we provide the policy with a privileged \textit{real goal image} of the fully completed task, eliding the need for any semantic understanding. Thus, the only advantage of \methodname\ over Oracle GCBC is the subgoal guidance, while \methodname\ is disadvantaged by additionally needing to interpret the language instruction and generate subgoals for the correct task.

As demonstrated in Table~\ref{tab:compare_to_oracle}, although Oracle GCBC is a strong baseline, \methodname\ is still the best-performing method on average. In particular, we observed that Oracle GCBC is still unable to grasp the bell pepper in Scene B due to a lack of low-level precision. Figure~\ref{fig:rollouts} provides a qualitative visualization of \methodname\ grasping the bell pepper while Oracle GCBC and RT-2-X fail to do so. These results validate the hypothesis that, in addition to enabling semantic generalization, \methodname\ also improves low-level precision and dexterity by decomposing the problem into a two-level hierarchy.

\begin{figure}[htbp]
    \centering
    \includegraphics[width=\linewidth]{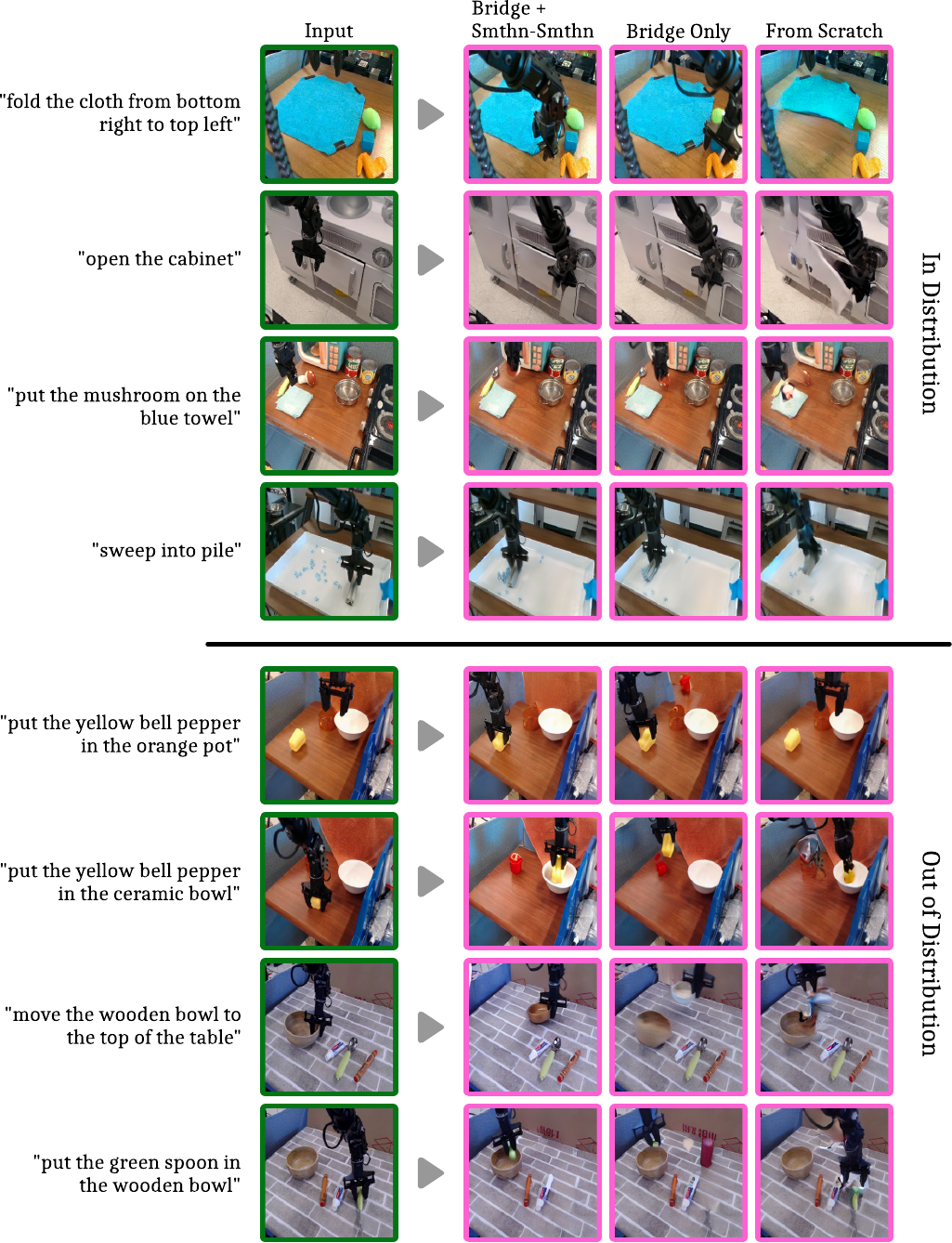}
    \caption{
        \figtitle{Comparison of subgoal quality} A comparison between 3 subgoal generation models: one trained on both robot and video data (Bridge + Smthn-Smthn), used throughout this paper; one trained on robot data only (Bridge Only); and one trained without InstructPix2Pix initialization (From Scratch). The top half (In Distribution) comes from the BridgeData V2 validation set, while the bottom half (Out of Distribution) comes from our evaluation Scenes B and C, unseen in BridgeData V2. Both Internet-scale pretraining and video cotraining and important for generating high-quality subgoals, especially in the zero-shot generalization setting.
    }
    \label{fig:sample_generations}
\end{figure}

\begin{table}[tbp]
    \centering
    \begin{tabular}{llcc}
        \toprule
        & Task & BridgeData Only & BridgeData + Something-Something \\
        \midrule
        \multirow{3}{*}[-3pt]{Scene B} & Bell pepper in pot & 0.2 & \textbf{0.5} \\
        & Bell pepper in bowl & 0.4 & \textbf{0.5} \\
        \cmidrule(lr){2-4}
        & Average & 0.30 & \textbf{0.50} \\
        \midrule
        \multirow{5}{*}[-3pt]{Scene C} & Toothpaste in bowl & \textbf{0.7} & 0.6 \\
        & Crayon in bowl & 0.9 & \textbf{1.0} \\
        & Spoon in bowl & \textbf{0.9} & \textbf{0.9}  \\
        & Bowl to top & 0.7 & \textbf{1.0}  \\
        \cmidrule(lr){2-4}
        & Average & 0.80 & \textbf{0.88} \\
        \bottomrule
    \end{tabular}
    \vspace{-0.25cm}
    \caption{\figtitle{Comparison to BridgeData-only subgoal generation model} Cotraining the subgoal generation model on the Something-Something dataset improves performance in Scenes B and C, which are unseen in BridgeData V2.}
    \label{tab:bridge_only}
    \vspace{-0.45cm}
\end{table}

\vspace{-0.3cm}
\subsection{Are Internet and Video Training Crucial for Zero-Shot Generalization?}
\vspace{-0.25cm}

Finally, we conduct an experiment to understand the effect of both Internet-scale pretraining (in the form of InstructPix2Pix initialization) and video cotraining (in the form of training on BridgeData V2 and Something-Something simultaneously). We train two additional subgoal generation models: one robot-only  model trained with InstructPix2Pix initialization but only on BridgeData V2, and one from-scratch model trained without InstructPix2Pix initialization at all. The from-scratch model uses the same frozen text encoder, image encoder, and UNet architecture as InstructPix2Pix, but initializes the UNet with random weights.

We present sample generations from these models in Figure~\ref{fig:sample_generations}. The generations of the from-scratch model are consistently low-quality, either misunderstanding the task, producing significant artifacts, or failing to edit the image at all; this demonstrates the importance of initializing from Internet-scale pretrained weights. As one might expect, the BridgeData-only model is on par with the full Something-Something model on in-distribution tasks, which come from the BridgeData V2 validation set. However, on out-of-distribution tasks, the BridgeData-only model has an increased propensity to hallucinate background items or misunderstand the task. We found that for moving the wooden bowl, in particular, the full Something-Something model is the only one that can produce fully correct subgoals.

As additional evidence, we present a quantitative evaluation of the BridgeData-only model on the tasks from Scenes B and C in Table~\ref{tab:bridge_only}. In real-world task execution, the difference between the BridgeData-only model and the full Something-Something model is not as pronounced as in the qualitative samples. This is because we find that the low-level goal-conditioned policy is fairly robust to hallucinations of background items, suggesting that it primarily attends to the gripper pose and the target object. However, the overall increased quality of the subgoals from the full model still produces a small quantitative improvement on average.

\vspace{-0.3cm}
\section{Discussion and Future Work}
\vspace{-0.3cm}

We presented a method for robotic control from language instructions that generates subgoals to guide a low-level goal-conditioned policy. The subgoals are generated by an image-editing diffusion model finetuned on video data.
This system improves both zero-shot generalization to new objects and the overall precision of the policy because the subgoal model incorporates semantic benefits from pretraining and commands the low-level policy with fine-grained guidance. Our experiments show that \methodname{} improves over prior techniques on the CALVIN benchmark and improves visual generalization and low-level control on real-world manipulation tasks. In the real world, \methodname\ outperforms language-conditioned behavioral cloning, a privileged goal-conditioned policy that gets access to a ground-truth final goal, as well the state-of-the-art instruction-following approach, RT-2-X, which is trained on more than an order of magnitude more robot data (over 1.1 million trajectories for RT-2-X, vs. 60k for ours).

Our method is simple and provides good performance, but it does have limitations that suggest promising directions for future work. 
For instance, the diffusion model and the low-level policy are trained separately, meaning the diffusion model is unaware of the capabilities of the low-level policy --- it is trained on the same data, but assumes that anything that is reachable in the data can also be reached by the policy. We hypothesize that performance can be improved by making the diffusion model aware of the low-level policy's capabilities.  More broadly, we found the performance of our method to often be bottlenecked by the performance of the low-level policy, suggesting that addressing either of these limitations might lead to a more performant method for importing Internet-scale knowledge into robotic manipulation.

\subsubsection*{Acknowledgments}
We thank Quan Vuong, Vincent Vanhoucke, Karl Pertsch, and Oleh Rybkin for their feedback on earlier versions of the paper. We thank Anurag Ajay, Zoey Chen, and Quan Vuong for their support with baseline approaches. Aviral Kumar thanks Abhishek Gupta, Sherry Yang, and Boyi Li for informative discussions. We also thank the TRC program from Google Cloud for their cloud TPU donations that were crucial for running our experiments.

Kevin Black and Pranav Atreya are supported by the NSF Graduate Research Fellowship. Mitsuhiko Nakamoto is supported by the Nakajima Foundation Fellowship.

\bibliography{main}

\begin{thebibliography}{65}
\providecommand{\natexlab}[1]{#1}
\providecommand{\url}[1]{\texttt{#1}}
\expandafter\ifx\csname urlstyle\endcsname\relax
  \providecommand{\doi}[1]{doi: #1}\else
  \providecommand{\doi}{doi: \begingroup \urlstyle{rm}\Url}\fi

\bibitem[Ajay et~al.(2023{\natexlab{a}})Ajay, Du, Gupta, Tenenbaum, Jaakkola,
  and Agrawal]{ajay2023is}
Anurag Ajay, Yilun Du, Abhi Gupta, Joshua~B. Tenenbaum, Tommi~S. Jaakkola, and
  Pulkit Agrawal.
\newblock Is conditional generative modeling all you need for decision making?
\newblock In \emph{The Eleventh International Conference on Learning
  Representations}, 2023{\natexlab{a}}.
\newblock URL \url{https://openreview.net/forum?id=sP1fo2K9DFG}.

\bibitem[Ajay et~al.(2023{\natexlab{b}})Ajay, Han, Du, Li, Gupta, Jaakkola,
  Tenenbaum, Kaelbling, Srivastava, and Agrawal]{hip}
Anurag Ajay, Seungwook Han, Yilun Du, Shaung Li, Abhi Gupta, Tommi Jaakkola,
  Josh Tenenbaum, Leslie Kaelbling, Akash Srivastava, and Pulkit Agrawal.
\newblock Compositional foundation models for hierarchical planning.
\newblock \emph{arXiv preprint arXiv:2309.08587}, 2023{\natexlab{b}}.

\bibitem[Babaeizadeh et~al.(2020)Babaeizadeh, Saffar, Nair, Levine, Finn, and
  Erhan]{babaeizadeh2021fitvid}
Mohammad Babaeizadeh, Mohammad~Taghi Saffar, Suraj Nair, Sergey Levine, Chelsea
  Finn, and Dumitru Erhan.
\newblock Fitvid: Overfitting in pixel-level video prediction.
\newblock \emph{arXiv preprint arXiv:2106.13195}, 2020.

\bibitem[Bhateja et~al.(2023)Bhateja, Guo, Ghosh, Singh, Tomar, Vuong,
  Chebotar, Levine, and Kumar]{vptr}
Chethan Bhateja, Derek Guo, Dibya Ghosh, Anika Singh, Manan Tomar, Quan~Ho
  Vuong, Yevgen Chebotar, Sergey Levine, and Aviral Kumar.
\newblock Robotic offline rl from internet videos via value-function
  pre-training.
\newblock 2023.
\newblock URL \url{https://api.semanticscholar.org/CorpusID:262217278}.

\bibitem[Brockman et~al.(2023)Brockman, Eleti, Georges, Jang, Kilpatrick, Lim,
  Miller, and Pokrass]{openai2023gpt3turbo}
Greg Brockman, Atty Eleti, Elie Georges, Joanne Jang, Logan Kilpatrick, Rachel
  Lim, Luke Miller, and Michelle Pokrass.
\newblock Introducing {ChatGPT} and {Whisper} {APIs}.
\newblock OpenAI Blog, 2023.
\newblock URL
  \url{https://openai.com/blog/introducing-chatgpt-and-whisper-apis}.

\bibitem[Brohan et~al.(2022)Brohan, Brown, Carbajal, Chebotar, Dabis, Finn,
  Gopalakrishnan, Hausman, Herzog, Hsu, et~al.]{brohan2022rt}
Anthony Brohan, Noah Brown, Justice Carbajal, Yevgen Chebotar, Joseph Dabis,
  Chelsea Finn, Keerthana Gopalakrishnan, Karol Hausman, Alex Herzog, Jasmine
  Hsu, et~al.
\newblock Rt-1: Robotics transformer for real-world control at scale.
\newblock \emph{arXiv preprint arXiv:2212.06817}, 2022.

\bibitem[Brohan et~al.(2023{\natexlab{a}})Brohan, Brown, Carbajal, Chebotar,
  Chen, Choromanski, Ding, Driess, Dubey, Finn, et~al.]{rt2}
Anthony Brohan, Noah Brown, Justice Carbajal, Yevgen Chebotar, Xi~Chen,
  Krzysztof Choromanski, Tianli Ding, Danny Driess, Avinava Dubey, Chelsea
  Finn, et~al.
\newblock Rt-2: Vision-language-action models transfer web knowledge to robotic
  control.
\newblock \emph{arXiv preprint arXiv:2307.15818}, 2023{\natexlab{a}}.

\bibitem[Brohan et~al.(2023{\natexlab{b}})Brohan, Chebotar, Finn, Hausman,
  Herzog, Ho, Ibarz, Irpan, Jang, Julian, et~al.]{saycan}
Anthony Brohan, Yevgen Chebotar, Chelsea Finn, Karol Hausman, Alexander Herzog,
  Daniel Ho, Julian Ibarz, Alex Irpan, Eric Jang, Ryan Julian, et~al.
\newblock Do as i can, not as i say: Grounding language in robotic affordances.
\newblock In \emph{Conference on Robot Learning}, pp.\  287--318. PMLR,
  2023{\natexlab{b}}.

\bibitem[Brooks et~al.(2023)Brooks, Holynski, and Efros]{instructpix2pix}
Tim Brooks, Aleksander Holynski, and Alexei~A Efros.
\newblock Instructpix2pix: Learning to follow image editing instructions.
\newblock In \emph{Proceedings of the IEEE/CVF Conference on Computer Vision
  and Pattern Recognition}, pp.\  18392--18402, 2023.

\bibitem[Chen et~al.(2021)Chen, Lu, Rajeswaran, Lee, Grover, Laskin, Abbeel,
  Srinivas, and Mordatch]{chen2021decision}
Lili Chen, Kevin Lu, Aravind Rajeswaran, Kimin Lee, Aditya Grover, Michael
  Laskin, Pieter Abbeel, Aravind Srinivas, and Igor Mordatch.
\newblock Decision transformer: Reinforcement learning via sequence modeling.
\newblock \emph{arXiv preprint arXiv:2106.01345}, 2021.

\bibitem[Chen et~al.(2023)Chen, Kiami, Gupta, and Kumar]{genaug}
Zoey Chen, Sho Kiami, Abhishek Gupta, and Vikash Kumar.
\newblock Genaug: Retargeting behaviors to unseen situations via generative
  augmentation.
\newblock \emph{arXiv preprint arXiv:2302.06671}, 2023.

\bibitem[Chi et~al.(2023)Chi, Feng, Du, Xu, Cousineau, Burchfiel, and
  Song]{chi2023diffusion}
Cheng Chi, Siyuan Feng, Yilun Du, Zhenjia Xu, Eric Cousineau, Benjamin
  Burchfiel, and Shuran Song.
\newblock Diffusion policy: Visuomotor policy learning via action diffusion.
\newblock \emph{arXiv preprint arXiv:2303.04137}, 2023.

\bibitem[Chung et~al.(2022)Chung, Hou, Longpre, Zoph, Tay, Fedus, Li, Wang,
  Dehghani, Brahma, et~al.]{flan-t5}
Hyung~Won Chung, Le~Hou, Shayne Longpre, Barret Zoph, Yi~Tay, William Fedus,
  Eric Li, Xuezhi Wang, Mostafa Dehghani, Siddhartha Brahma, et~al.
\newblock Scaling instruction-finetuned language models.
\newblock \emph{arXiv preprint arXiv:2210.11416}, 2022.

\bibitem[Collaboration et~al.(2023)Collaboration, Padalkar, Pooley, Jain,
  Bewley, Herzog, Irpan, Khazatsky, Rai, Singh, Brohan, Raffin, Wahid,
  Burgess-Limerick, Kim, Schölkopf, Ichter, Lu, Xu, Finn, Xu, Chi, Huang,
  Chan, Pan, Fu, Devin, Driess, Pathak, Shah, Büchler, Kalashnikov, Sadigh,
  Johns, Ceola, Xia, Stulp, Zhou, Sukhatme, Salhotra, Yan, Schiavi, Su, Fang,
  Shi, Amor, Christensen, Furuta, Walke, Fang, Mordatch, Radosavovic, Leal,
  Liang, Kim, Schneider, Hsu, Bohg, Bingham, Wu, Wu, Luo, Gu, Tan, Oh, Malik,
  Tompson, Yang, Lim, Silvério, Han, Rao, Pertsch, Hausman, Go,
  Gopalakrishnan, Goldberg, Byrne, Oslund, Kawaharazuka, Zhang, Majd, Rana,
  Srinivasan, Chen, Pinto, Tan, Ott, Lee, Tomizuka, Du, Ahn, Zhang, Ding,
  Srirama, Sharma, Kim, Kanazawa, Hansen, Heess, Joshi, Suenderhauf, Palo,
  Shafiullah, Mees, Kroemer, Sanketi, Wohlhart, Xu, Sermanet, Sundaresan,
  Vuong, Rafailov, Tian, Doshi, Martín-Martín, Mendonca, Shah, Hoque, Julian,
  Bustamante, Kirmani, Levine, Moore, Bahl, Dass, Song, Xu, Haldar, Adebola,
  Guist, Nasiriany, Schaal, Welker, Tian, Dasari, Belkhale, Osa, Harada,
  Matsushima, Xiao, Yu, Ding, Davchev, Zhao, Armstrong, Darrell, Jain,
  Vanhoucke, Zhan, Zhou, Burgard, Chen, Wang, Zhu, Li, Lu, Chebotar, Zhou, Zhu,
  Xu, Wang, Bisk, Cho, Lee, Cui, hua Wu, Tang, Zhu, Li, Iwasawa, Matsuo, Xu,
  and Cui]{rt2x}
Open X-Embodiment Collaboration, Abhishek Padalkar, Acorn Pooley, Ajinkya Jain,
  Alex Bewley, Alex Herzog, Alex Irpan, Alexander Khazatsky, Anant Rai, Anikait
  Singh, Anthony Brohan, Antonin Raffin, Ayzaan Wahid, Ben Burgess-Limerick,
  Beomjoon Kim, Bernhard Schölkopf, Brian Ichter, Cewu Lu, Charles Xu, Chelsea
  Finn, Chenfeng Xu, Cheng Chi, Chenguang Huang, Christine Chan, Chuer Pan,
  Chuyuan Fu, Coline Devin, Danny Driess, Deepak Pathak, Dhruv Shah, Dieter
  Büchler, Dmitry Kalashnikov, Dorsa Sadigh, Edward Johns, Federico Ceola, Fei
  Xia, Freek Stulp, Gaoyue Zhou, Gaurav~S. Sukhatme, Gautam Salhotra, Ge~Yan,
  Giulio Schiavi, Hao Su, Hao-Shu Fang, Haochen Shi, Heni~Ben Amor, Henrik~I
  Christensen, Hiroki Furuta, Homer Walke, Hongjie Fang, Igor Mordatch, Ilija
  Radosavovic, Isabel Leal, Jacky Liang, Jaehyung Kim, Jan Schneider, Jasmine
  Hsu, Jeannette Bohg, Jeffrey Bingham, Jiajun Wu, Jialin Wu, Jianlan Luo,
  Jiayuan Gu, Jie Tan, Jihoon Oh, Jitendra Malik, Jonathan Tompson, Jonathan
  Yang, Joseph~J. Lim, João Silvério, Junhyek Han, Kanishka Rao, Karl
  Pertsch, Karol Hausman, Keegan Go, Keerthana Gopalakrishnan, Ken Goldberg,
  Kendra Byrne, Kenneth Oslund, Kento Kawaharazuka, Kevin Zhang, Keyvan Majd,
  Krishan Rana, Krishnan Srinivasan, Lawrence~Yunliang Chen, Lerrel Pinto, Liam
  Tan, Lionel Ott, Lisa Lee, Masayoshi Tomizuka, Maximilian Du, Michael Ahn,
  Mingtong Zhang, Mingyu Ding, Mohan~Kumar Srirama, Mohit Sharma, Moo~Jin Kim,
  Naoaki Kanazawa, Nicklas Hansen, Nicolas Heess, Nikhil~J Joshi, Niko
  Suenderhauf, Norman~Di Palo, Nur Muhammad~Mahi Shafiullah, Oier Mees, Oliver
  Kroemer, Pannag~R Sanketi, Paul Wohlhart, Peng Xu, Pierre Sermanet, Priya
  Sundaresan, Quan Vuong, Rafael Rafailov, Ran Tian, Ria Doshi, Roberto
  Martín-Martín, Russell Mendonca, Rutav Shah, Ryan Hoque, Ryan Julian,
  Samuel Bustamante, Sean Kirmani, Sergey Levine, Sherry Moore, Shikhar Bahl,
  Shivin Dass, Shuran Song, Sichun Xu, Siddhant Haldar, Simeon Adebola, Simon
  Guist, Soroush Nasiriany, Stefan Schaal, Stefan Welker, Stephen Tian, Sudeep
  Dasari, Suneel Belkhale, Takayuki Osa, Tatsuya Harada, Tatsuya Matsushima,
  Ted Xiao, Tianhe Yu, Tianli Ding, Todor Davchev, Tony~Z. Zhao, Travis
  Armstrong, Trevor Darrell, Vidhi Jain, Vincent Vanhoucke, Wei Zhan, Wenxuan
  Zhou, Wolfram Burgard, Xi~Chen, Xiaolong Wang, Xinghao Zhu, Xuanlin Li, Yao
  Lu, Yevgen Chebotar, Yifan Zhou, Yifeng Zhu, Ying Xu, Yixuan Wang, Yonatan
  Bisk, Yoonyoung Cho, Youngwoon Lee, Yuchen Cui, Yueh hua Wu, Yujin Tang, Yuke
  Zhu, Yunzhu Li, Yusuke Iwasawa, Yutaka Matsuo, Zhuo Xu, and Zichen~Jeff Cui.
\newblock Open {X-E}mbodiment: Robotic learning datasets and {RT-X} models.
\newblock \url{https://robotics-transformer-x.github.io}, 2023.

\bibitem[Driess et~al.(2023)Driess, Xia, Sajjadi, Lynch, Chowdhery, Ichter,
  Wahid, Tompson, Vuong, Yu, et~al.]{palme}
Danny Driess, Fei Xia, Mehdi~SM Sajjadi, Corey Lynch, Aakanksha Chowdhery,
  Brian Ichter, Ayzaan Wahid, Jonathan Tompson, Quan Vuong, Tianhe Yu, et~al.
\newblock Palm-e: An embodied multimodal language model.
\newblock \emph{arXiv preprint arXiv:2303.03378}, 2023.

\bibitem[Du et~al.(2023)Du, Yang, Dai, Dai, Nachum, Tenenbaum, Schuurmans, and
  Abbeel]{unipi}
Yilun Du, Mengjiao Yang, Bo~Dai, Hanjun Dai, Ofir Nachum, Joshua~B Tenenbaum,
  Dale Schuurmans, and Pieter Abbeel.
\newblock Learning universal policies via text-guided video generation.
\newblock \emph{arXiv preprint arXiv:2302.00111}, 2023.

\bibitem[Ebert et~al.(2018)Ebert, Finn, Dasari, Xie, Lee, and
  Levine]{ebert2018visual}
Frederik Ebert, Chelsea Finn, Sudeep Dasari, Annie Xie, Alex Lee, and Sergey
  Levine.
\newblock Visual foresight: Model-based deep reinforcement learning for
  vision-based robotic control.
\newblock \emph{arXiv preprint arXiv:1812.00568}, 2018.

\bibitem[Ge et~al.(2023)Ge, Macaluso, Li, Luo, and Wang]{ge2023policy}
Yuying Ge, Annabella Macaluso, Li~Erran Li, Ping Luo, and Xiaolong Wang.
\newblock Policy adaptation from foundation model feedback.
\newblock In \emph{Proceedings of the IEEE/CVF Conference on Computer Vision
  and Pattern Recognition}, pp.\  19059--19069, 2023.

\bibitem[Goyal et~al.(2017)Goyal, Ebrahimi~Kahou, Michalski, Materzynska,
  Westphal, Kim, Haenel, Fruend, Yianilos, Mueller-Freitag,
  et~al.]{goyal2017something}
Raghav Goyal, Samira Ebrahimi~Kahou, Vincent Michalski, Joanna Materzynska,
  Susanne Westphal, Heuna Kim, Valentin Haenel, Ingo Fruend, Peter Yianilos,
  Moritz Mueller-Freitag, et~al.
\newblock The" something something" video database for learning and evaluating
  visual common sense.
\newblock In \emph{Proceedings of the IEEE international conference on computer
  vision}, pp.\  5842--5850, 2017.

\bibitem[Grauman et~al.(2022)Grauman, Westbury, Byrne, Chavis, Furnari,
  Girdhar, Hamburger, Jiang, Liu, Liu, et~al.]{grauman2022ego4d}
Kristen Grauman, Andrew Westbury, Eugene Byrne, Zachary Chavis, Antonino
  Furnari, Rohit Girdhar, Jackson Hamburger, Hao Jiang, Miao Liu, Xingyu Liu,
  et~al.
\newblock Ego4d: Around the world in 3,000 hours of egocentric video.
\newblock In \emph{Proceedings of the IEEE/CVF Conference on Computer Vision
  and Pattern Recognition}, pp.\  18995--19012, 2022.

\bibitem[Hafner et~al.(2019)Hafner, Lillicrap, Ba, and
  Norouzi]{hafner2019dream}
Danijar Hafner, Timothy Lillicrap, Jimmy Ba, and Mohammad Norouzi.
\newblock Dream to control: Learning behaviors by latent imagination.
\newblock \emph{arXiv preprint arXiv:1912.01603}, 2019.

\bibitem[Hafner et~al.(2023)Hafner, Pasukonis, Ba, and
  Lillicrap]{hafner2023mastering}
Danijar Hafner, Jurgis Pasukonis, Jimmy Ba, and Timothy Lillicrap.
\newblock Mastering diverse domains through world models.
\newblock \emph{arXiv preprint arXiv:2301.04104}, 2023.

\bibitem[Hansen-Estruch et~al.(2023)Hansen-Estruch, Kostrikov, Janner, Kuba,
  and Levine]{hansen2023idql}
Philippe Hansen-Estruch, Ilya Kostrikov, Michael Janner, Jakub~Grudzien Kuba,
  and Sergey Levine.
\newblock Idql: Implicit q-learning as an actor-critic method with diffusion
  policies.
\newblock \emph{arXiv preprint arXiv:2304.10573}, 2023.

\bibitem[Ho \& Salimans(2022)Ho and Salimans]{classifier-free}
Jonathan Ho and Tim Salimans.
\newblock Classifier-free diffusion guidance.
\newblock \emph{arXiv preprint arXiv:2207.12598}, 2022.

\bibitem[Ho et~al.(2020)Ho, Jain, and Abbeel]{ho2020denoising}
Jonathan Ho, Ajay Jain, and Pieter Abbeel.
\newblock Denoising diffusion probabilistic models.
\newblock \emph{Advances in Neural Information Processing Systems},
  33:\penalty0 6840--6851, 2020.

\bibitem[Ho et~al.(2022{\natexlab{a}})Ho, Chan, Saharia, Whang, Gao, Gritsenko,
  Kingma, Poole, Norouzi, Fleet, and Salimans]{imagen-video}
Jonathan Ho, William Chan, Chitwan Saharia, Jay Whang, Ruiqi Gao, Alexey
  Gritsenko, Diederik~P. Kingma, Ben Poole, Mohammad Norouzi, David~J. Fleet,
  and Tim Salimans.
\newblock Imagen video: High definition video generation with diffusion models,
  2022{\natexlab{a}}.

\bibitem[Ho et~al.(2022{\natexlab{b}})Ho, Salimans, Gritsenko, Chan, Norouzi,
  and Fleet]{vdm}
Jonathan Ho, Tim Salimans, Alexey Gritsenko, William Chan, Mohammad Norouzi,
  and David~J Fleet.
\newblock Video diffusion models.
\newblock \emph{arXiv:2204.03458}, 2022{\natexlab{b}}.

\bibitem[Huang et~al.(2022{\natexlab{a}})Huang, Abbeel, Pathak, and
  Mordatch]{huang2022language}
Wenlong Huang, Pieter Abbeel, Deepak Pathak, and Igor Mordatch.
\newblock Language models as zero-shot planners: Extracting actionable
  knowledge for embodied agents.
\newblock \emph{arXiv preprint arXiv:2201.07207}, 2022{\natexlab{a}}.

\bibitem[Huang et~al.(2022{\natexlab{b}})Huang, Xia, Xiao, Chan, Liang,
  Florence, Zeng, Tompson, Mordatch, Chebotar, et~al.]{huang2022inner}
Wenlong Huang, Fei Xia, Ted Xiao, Harris Chan, Jacky Liang, Pete Florence, Andy
  Zeng, Jonathan Tompson, Igor Mordatch, Yevgen Chebotar, et~al.
\newblock Inner monologue: Embodied reasoning through planning with language
  models.
\newblock \emph{arXiv preprint arXiv:2207.05608}, 2022{\natexlab{b}}.

\bibitem[Jang et~al.(2022)Jang, Irpan, Khansari, Kappler, Ebert, Lynch, Levine,
  and Finn]{jang2022bc}
Eric Jang, Alex Irpan, Mohi Khansari, Daniel Kappler, Frederik Ebert, Corey
  Lynch, Sergey Levine, and Chelsea Finn.
\newblock Bc-z: Zero-shot task generalization with robotic imitation learning.
\newblock In \emph{Conference on Robot Learning}, pp.\  991--1002. PMLR, 2022.

\bibitem[Janner et~al.(2021)Janner, Li, and Levine]{janner2021sequence}
Michael Janner, Qiyang Li, and Sergey Levine.
\newblock Offline reinforcement learning as one big sequence modeling problem.
\newblock In \emph{Advances in Neural Information Processing Systems}, 2021.

\bibitem[Janner et~al.(2022)Janner, Du, Tenenbaum, and
  Levine]{janner2022diffuser}
Michael Janner, Yilun Du, Joshua Tenenbaum, and Sergey Levine.
\newblock Planning with diffusion for flexible behavior synthesis.
\newblock In \emph{International Conference on Machine Learning}, 2022.

\bibitem[Kamath et~al.(2021)Kamath, Singh, LeCun, Synnaeve, Misra, and
  Carion]{kamath2021mdetr}
Aishwarya Kamath, Mannat Singh, Yann LeCun, Gabriel Synnaeve, Ishan Misra, and
  Nicolas Carion.
\newblock Mdetr-modulated detection for end-to-end multi-modal understanding.
\newblock In \emph{Proceedings of the IEEE/CVF International Conference on
  Computer Vision}, pp.\  1780--1790, 2021.

\bibitem[Kapelyukh et~al.(2023)Kapelyukh, Vosylius, and Johns]{dall-e-bot}
Ivan Kapelyukh, Vitalis Vosylius, and Edward Johns.
\newblock Dall-e-bot: Introducing web-scale diffusion models to robotics.
\newblock 2023.

\bibitem[Karamcheti et~al.(2023)Karamcheti, Nair, Chen, Kollar, Finn, Sadigh,
  and Liang]{voltron}
Siddharth Karamcheti, Suraj Nair, Annie~S Chen, Thomas Kollar, Chelsea Finn,
  Dorsa Sadigh, and Percy Liang.
\newblock Language-driven representation learning for robotics.
\newblock \emph{arXiv preprint arXiv:2302.12766}, 2023.

\bibitem[Kingma \& Ba(2015)Kingma and Ba]{kingma2014adam}
Diederik Kingma and Jimmy Ba.
\newblock {Adam: A method for stochastic optimization}.
\newblock \emph{International Conference on Learning Representations (ICLR)},
  2015.

\bibitem[Lee et~al.(2020)Lee, Nagabandi, Abbeel, and Levine]{lee2020stochastic}
Alex~X Lee, Anusha Nagabandi, Pieter Abbeel, and Sergey Levine.
\newblock Stochastic latent actor-critic: Deep reinforcement learning with a
  latent variable model.
\newblock \emph{Advances in Neural Information Processing Systems},
  33:\penalty0 741--752, 2020.

\bibitem[Lee \& He(2018)Lee and He]{lee2018stochastic}
Donghwan Lee and Niao He.
\newblock Stochastic primal-dual q-learning.
\newblock \emph{arXiv preprint arXiv:1810.08298}, 2018.

\bibitem[Liang et~al.(2023)Liang, Huang, Xia, Xu, Hausman, Ichter, Florence,
  and Zeng]{liang2023code}
Jacky Liang, Wenlong Huang, Fei Xia, Peng Xu, Karol Hausman, Brian Ichter, Pete
  Florence, and Andy Zeng.
\newblock Code as policies: Language model programs for embodied control.
\newblock In \emph{2023 IEEE International Conference on Robotics and
  Automation (ICRA)}, pp.\  9493--9500. IEEE, 2023.

\bibitem[Loshchilov \& Hutter(2017)Loshchilov and
  Hutter]{loshchilov2017decoupled}
Ilya Loshchilov and Frank Hutter.
\newblock Decoupled weight decay regularization.
\newblock \emph{arXiv preprint arXiv:1711.05101}, 2017.

\bibitem[Lynch \& Sermanet(2020)Lynch and Sermanet]{lynch2020language}
Corey Lynch and Pierre Sermanet.
\newblock Language conditioned imitation learning over unstructured data.
\newblock \emph{arXiv preprint arXiv:2005.07648}, 2020.

\bibitem[Ma et~al.(2023)Ma, Liang, Som, Kumar, Zhang, Bastani, and
  Jayaraman]{liv}
Yecheng~Jason Ma, William Liang, Vaidehi Som, Vikash Kumar, Amy Zhang, Osbert
  Bastani, and Dinesh Jayaraman.
\newblock Liv: Language-image representations and rewards for robotic control.
\newblock \emph{arXiv preprint arXiv:2306.00958}, 2023.

\bibitem[Mandi et~al.(2022)Mandi, Bharadhwaj, Moens, Song, Rajeswaran, and
  Kumar]{cacti}
Zhao Mandi, Homanga Bharadhwaj, Vincent Moens, Shuran Song, Aravind Rajeswaran,
  and Vikash Kumar.
\newblock Cacti: A framework for scalable multi-task multi-scene visual
  imitation learning.
\newblock \emph{arXiv preprint arXiv:2212.05711}, 2022.

\bibitem[Mees et~al.(2022{\natexlab{a}})Mees, Hermann, and Burgard]{hulc}
Oier Mees, Lukas Hermann, and Wolfram Burgard.
\newblock What matters in language conditioned robotic imitation learning over
  unstructured data.
\newblock \emph{IEEE Robotics and Automation Letters}, 7\penalty0 (4):\penalty0
  11205--11212, 2022{\natexlab{a}}.

\bibitem[Mees et~al.(2022{\natexlab{b}})Mees, Hermann, Rosete-Beas, and
  Burgard]{calvin}
Oier Mees, Lukas Hermann, Erick Rosete-Beas, and Wolfram Burgard.
\newblock Calvin: A benchmark for language-conditioned policy learning for
  long-horizon robot manipulation tasks.
\newblock \emph{IEEE Robotics and Automation Letters (RA-L)}, 7\penalty0
  (3):\penalty0 7327--7334, 2022{\natexlab{b}}.

\bibitem[Minderer et~al.(2022)Minderer, Gritsenko, Stone, Neumann, Weissenborn,
  Dosovitskiy, Mahendran, Arnab, Dehghani, Shen, et~al.]{owl-vit}
Matthias Minderer, Alexey Gritsenko, Austin Stone, Maxim Neumann, Dirk
  Weissenborn, Alexey Dosovitskiy, Aravindh Mahendran, Anurag Arnab, Mostafa
  Dehghani, Zhuoran Shen, et~al.
\newblock Simple open-vocabulary object detection.
\newblock In \emph{European Conference on Computer Vision}, pp.\  728--755.
  Springer, 2022.

\bibitem[Myers et~al.(2023)Myers, He, Fang, Walke, Hansen-Estruch, Cheng,
  Jalobeanu, Kolobov, Dragan, and Levine]{grif}
Vivek Myers, Andre He, Kuan Fang, Homer Walke, Philippe Hansen-Estruch,
  Ching-An Cheng, Mihai Jalobeanu, Andrey Kolobov, Anca Dragan, and Sergey
  Levine.
\newblock Goal representations for instruction following: A semi-supervised
  language interface to control.
\newblock \emph{arXiv preprint arXiv:2307.00117}, 2023.

\bibitem[Nair \& Finn(2019)Nair and Finn]{nair2019hierarchical}
Suraj Nair and Chelsea Finn.
\newblock Hierarchical foresight: Self-supervised learning of long-horizon
  tasks via visual subgoal generation.
\newblock \emph{arXiv preprint arXiv:1909.05829}, 2019.

\bibitem[Nair et~al.(2022)Nair, Rajeswaran, Kumar, Finn, and Gupta]{r3m}
Suraj Nair, Aravind Rajeswaran, Vikash Kumar, Chelsea Finn, and Abhi Gupta.
\newblock R3m: A universal visual representation for robot manipulation.
\newblock \emph{ArXiv}, abs/2203.12601, 2022.

\bibitem[Pashevich et~al.(2019)Pashevich, Strudel, Kalevatykh, Laptev, and
  Schmid]{pashevich2019learning}
Alexander Pashevich, Robin Strudel, Igor Kalevatykh, Ivan Laptev, and Cordelia
  Schmid.
\newblock Learning to augment synthetic images for sim2real policy transfer.
\newblock In \emph{2019 IEEE/RSJ International Conference on Intelligent Robots
  and Systems (IROS)}, pp.\  2651--2657. IEEE, 2019.

\bibitem[Perez et~al.(2017)Perez, Strub, {de Vries}, Dumoulin, and
  Courville]{perezFiLMVisualReasoning2017}
Ethan Perez, Florian Strub, Harm {de Vries}, Vincent Dumoulin, and Aaron
  Courville.
\newblock {{FiLM}}: {{Visual Reasoning}} with a {{General Conditioning Layer}},
  December 2017.

\bibitem[Radford et~al.(2021)Radford, Kim, Hallacy, Ramesh, Goh, Agarwal,
  Sastry, Askell, Mishkin, Clark, et~al.]{clip}
Alec Radford, Jong~Wook Kim, Chris Hallacy, Aditya Ramesh, Gabriel Goh,
  Sandhini Agarwal, Girish Sastry, Amanda Askell, Pamela Mishkin, Jack Clark,
  et~al.
\newblock Learning transferable visual models from natural language
  supervision.
\newblock In \emph{International conference on machine learning}, pp.\
  8748--8763. PMLR, 2021.

\bibitem[Rafailov et~al.(2021)Rafailov, Yu, Rajeswaran, and
  Finn]{Rafailov2020LOMPO}
Rafael Rafailov, Tianhe Yu, A.~Rajeswaran, and Chelsea Finn.
\newblock Offline reinforcement learning from images with latent space models.
\newblock \emph{Learning for Decision Making and Control (L4DC)}, 2021.

\bibitem[Seo et~al.(2022)Seo, Lee, James, and Abbeel]{seo2022reinforcement}
Younggyo Seo, Kimin Lee, Stephen~L James, and Pieter Abbeel.
\newblock Reinforcement learning with action-free pre-training from videos.
\newblock In \emph{International Conference on Machine Learning}, pp.\
  19561--19579. PMLR, 2022.

\bibitem[Shridhar et~al.(2022)Shridhar, Manuelli, and Fox]{cliport}
Mohit Shridhar, Lucas Manuelli, and Dieter Fox.
\newblock Cliport: What and where pathways for robotic manipulation.
\newblock In \emph{Conference on Robot Learning}, pp.\  894--906. PMLR, 2022.

\bibitem[Song et~al.(2020)Song, Meng, and Ermon]{song2020denoising}
Jiaming Song, Chenlin Meng, and Stefano Ermon.
\newblock Denoising diffusion implicit models.
\newblock \emph{arXiv preprint arXiv:2010.02502}, 2020.

\bibitem[Stone et~al.(2023)Stone, Xiao, Lu, Gopalakrishnan, Lee, Vuong,
  Wohlhart, Zitkovich, Xia, Finn, et~al.]{moo}
Austin Stone, Ted Xiao, Yao Lu, Keerthana Gopalakrishnan, Kuang-Huei Lee, Quan
  Vuong, Paul Wohlhart, Brianna Zitkovich, Fei Xia, Chelsea Finn, et~al.
\newblock Open-world object manipulation using pre-trained vision-language
  models.
\newblock \emph{arXiv preprint arXiv:2303.00905}, 2023.

\bibitem[Villegas et~al.(2019)Villegas, Pathak, Kannan, Erhan, Le, and
  Lee]{villegas2019high}
Ruben Villegas, Arkanath Pathak, Harini Kannan, Dumitru Erhan, Quoc~V Le, and
  Honglak Lee.
\newblock High fidelity video prediction with large stochastic recurrent neural
  networks.
\newblock \emph{Advances in Neural Information Processing Systems}, 32, 2019.

\bibitem[Walke et~al.(2023)Walke, Black, Lee, Kim, Du, Zheng, Zhao,
  Hansen-Estruch, Vuong, He, Myers, Fang, Finn, and Levine]{bridgev2}
Homer Walke, Kevin Black, Abraham Lee, Moo~Jin Kim, Max Du, Chongyi Zheng, Tony
  Zhao, Philippe Hansen-Estruch, Quan Vuong, Andre He, Vivek Myers, Kuan Fang,
  Chelsea Finn, and Sergey Levine.
\newblock Bridgedata v2: A dataset for robot learning at scale.
\newblock In \emph{Conference on Robot Learning (CoRL)}, 2023.

\bibitem[Wu et~al.(2023)Wu, Escontrela, Hafner, Abbeel, and
  Goldberg]{wu2023daydreamer}
Philipp Wu, Alejandro Escontrela, Danijar Hafner, Pieter Abbeel, and Ken
  Goldberg.
\newblock Daydreamer: World models for physical robot learning.
\newblock In \emph{Conference on Robot Learning}, pp.\  2226--2240. PMLR, 2023.

\bibitem[Yang et~al.(2019)Yang, Cer, Ahmad, Guo, Law, Constant, Abrego, Yuan,
  Tar, Sung, et~al.]{yangMultilingualUniversalSentence2019}
Yinfei Yang, Daniel Cer, Amin Ahmad, Mandy Guo, Jax Law, Noah Constant,
  Gustavo~Hernandez Abrego, Steve Yuan, Chris Tar, Yun-Hsuan Sung, et~al.
\newblock Multilingual {{Universal Sentence Encoder}} for {{Semantic
  Retrieval}}, July 2019.

\bibitem[Yu et~al.(2023{\natexlab{a}})Yu, Sohn, Kim, and Shin]{pvdm}
Sihyun Yu, Kihyuk Sohn, Subin Kim, and Jinwoo Shin.
\newblock Video probabilistic diffusion models in projected latent space.
\newblock In \emph{Proceedings of the IEEE/CVF Conference on Computer Vision
  and Pattern Recognition}, pp.\  18456--18466, 2023{\natexlab{a}}.

\bibitem[Yu et~al.(2021)Yu, Kumar, Rafailov, Rajeswaran, Levine, and
  Finn]{yu2021combo}
Tianhe Yu, Aviral Kumar, Rafael Rafailov, Aravind Rajeswaran, Sergey Levine,
  and Chelsea Finn.
\newblock Combo: Conservative offline model-based policy optimization.
\newblock \emph{arXiv preprint arXiv:2102.08363}, 2021.

\bibitem[Yu et~al.(2023{\natexlab{b}})Yu, Xiao, Stone, Tompson, Brohan, Wang,
  Singh, Tan, Peralta, Ichter, et~al.]{rosie}
Tianhe Yu, Ted Xiao, Austin Stone, Jonathan Tompson, Anthony Brohan, Su~Wang,
  Jaspiar Singh, Clayton Tan, Jodilyn Peralta, Brian Ichter, et~al.
\newblock Scaling robot learning with semantically imagined experience.
\newblock \emph{arXiv preprint arXiv:2302.11550}, 2023{\natexlab{b}}.

\bibitem[Zhao et~al.(2023)Zhao, Kumar, Levine, and Finn]{zhao2023learning}
Tony~Z Zhao, Vikash Kumar, Sergey Levine, and Chelsea Finn.
\newblock Learning fine-grained bimanual manipulation with low-cost hardware.
\newblock \emph{arXiv preprint arXiv:2304.13705}, 2023.

\end{thebibliography}
\bibliographystyle{setup/iclr2024_conference}

\newpage

\appendix
\section{Implementation Details}

We provide implementation details for \methodname{} and the baselines. Table~\ref{tab:hyperparameters} documents goal sampling hyperparameters.

\subsection{\methodname{} implementation details}
\label{appendix:ours}

\begin{table}
    \centering
    \begin{tabular}{rccc}
        \toprule
        & CALVIN & Smthn-Smthn & Bridge \\
        \midrule
        $k_\text{min}$ (Eq.~\ref{eqn:finetuning_phase1}) & 20 & 11 & 11 \\
        $k_\text{max}$ (Eq.~\ref{eqn:finetuning_phase1}) & 22 & 14 & 14 \\
        $k_\delta$ (Eq.~\ref{eqn:gcbc_training}) & 2 & N/A & 6 \\
        \bottomrule
    \end{tabular} \\[1em]
    \begin{tabular}{lcc}
        \toprule
        & CALVIN & Real-world \\
        \midrule
        $k_\text{test}$ (Alg.~\ref{algo:planning}) & 20 & 20 \\
        \bottomrule
    \end{tabular}
    \caption{\figtitle{Goal sampling hyperparameters} The first two hyperparameters, $k_\text{min}$ and $k_\text{max}$, are used for training the image-editing model. $k_\delta$ is used for training the low-level policy. All 3 of these correspond to a particular dataset.
    $k_\text{test}$ is a test-time hyperparameter that corresponds to an evaluation setup.}
    \label{tab:hyperparameters}
\end{table}

\subsubsection{Image-editing diffusion model}
\label{appendix:ours-image}

We finetune InstructPix2Pix~\citep{instructpix2pix} using similar hyperparameters to the initial InstructPix2Pix training. We use the AdamW optimizer~\citep{loshchilov2017decoupled} with a learning rate of 1e-4, a linear warmup of 800 steps, and weight decay of 0.01. We track an exponential moving average (EMA) of the model parameters with a decay rate of 0.999 and use the EMA parameters at test time. We train for 40k steps with a batch size of 1024 on a single v4-64 TPU pod, which takes 17 hours.

For simulation, we train the model exclusively on the CALVIN dataset. For real-world experiments, we cotrain with a sampling mixture of 60\% Something-Something and 40\% BridgeData V2.

At test time, we use an image guidance weight of 2.5 and a text guidance weight of 7.5. We use the DDIM sampler~\citep{song2020denoising} with 50 sampling steps.

\subsubsection{Goal-reaching policy}
\label{appendix:ours-policy}

We use a diffusion model for our goal-reaching policy since recent work has shown that diffusion-based policies can better capture multi-modality in robot data \citep{chi2023diffusion, hansen2023idql}, leading to improved performance across a variety of tasks. In our implementation (which follows \citet{bridgev2}), the observation and goal image are stacked channel-wise before being passed into a ResNet-50 image encoder. This image encoding is used to condition a diffusion process that models the action distribution. The diffusion process uses an MLP with 3 256-unit layers and residual connections. Following \citet{chi2023diffusion}, rather than predicting a single action, we predict a sequence of 4 actions to encourage temporal consistency. We use the Adam optimizer \citep{kingma2014adam} with a learning rate of 3e-4 and a linear warmup of 2000 steps. We train with a batch size of 256 for 445k steps on a single v4-8 TPU VM, which takes 15 hours. We augment the observation and goal with random crops, random resizing, and color jitter. For goal sampling, we use $k_\delta = 6$, meaning we sample goals uniformly from the range $[0, 20)$, since the corresponding $k_\text{max}$ used with BridgeData V2 is 14 (see Table~\ref{tab:hyperparameters}).

At test time, we have several options for how to predict and execute action sequences. \citet{chi2023diffusion} use receding horizon control, sampling $k$-length action sequences and only executing some of the actions before sampling a new sequence. This strategy can make the policy more reactive. However, we found that the robot behavior was quite jerky as the policy switched between different modes in the action distribution with each sample. Instead, we use a temporal ensembling strategy similar to \citet{zhao2023learning}. We predict a new 4-action sequence at each timestep and execute the dimension-wise mean of the last 4 predictions for that timestep.

\subsection{Baseline implementation details}
\label{appendix:baselines}

\subsubsection{MOO}
\label{appendix:moo}
MOO~\citep{moo} utilizes a mask to represent the target objects and incorporates it as an additional channel in the observation. Specifically, they train a language-conditioned policy that takes a 4-channel image and a language command as inputs. To acquire the mask for target objects, the OWL-ViT~\citep{owl-vit} object detector is employed. This detector is an open-vocabulary object detection model, pretrained on Internet-scale datasets, and it is used to extract the bounding boxes of the objects of interest from the image. For tasks like ``move X to Y,'' MOO extracts the bounding box for X, representing the object of interest, and Y, indicating the target place. A mask is then created where the pixel at the center of the predicted bounding box is assigned a value of 1.0 for X and 0.5 for Y.

\textbf{Extracting object entities from BridgeData V2 language annotations.} In order to obtain the mask, it is necessary to extract the entities corresponding to the object of interest, denoted as X, and the target place, Y, from the language command. In the MOO paper, the authors assume that the language in their dataset is structured in a way that facilitates easy separation of X and Y. Specifically, they employ a dataset that exclusively consists of language annotations such as ``pick X,'' ``move X near Y,'' ``knock X over,'' ``place X upright,'' and ``place X into Y.'' 

Given that the language annotations in BridgeData V2 are diverse and unstructured, it is challenging to naively extract X and Y. We utilized OpenAI's \texttt{gpt-3.5-turbo-instruct} model~\citep{openai2023gpt3turbo} to extract the object of interest and the target place (if any) from the language annotation and input them into OWL-ViT to create masks. It is worth noting that due to the unstructured nature of BridgeData V2, the extracted bounding boxes in training data are sometimes incorrect as shown in Figure~\ref{fig:moo_mask}. We then trained a mask-conditioned LCBC policy using the same architecture as described in Appendix~\ref{appendix:lcbc}. Following MOO, we removed X and Y from the prompt and replaced the entity X with ``object of interest'' and the entity Y with ``target place''. For example, given the language prompt ``put the eggplant in the pot'', we use a modified prompt ``put object of interest in target place'' as the input to the policy during both train and test time.

\textbf{Test time.} During test time, we manually give ground-truth masks to the policy. To enable this, we build a simple interface on the robot machine, allowing the evaluator to create the masks by clicking on the initial camera image at the beginning of each trial.
\begin{figure}[h]
    \centering
    \vspace{-0.2cm}
    \includegraphics[width=0.9\linewidth]{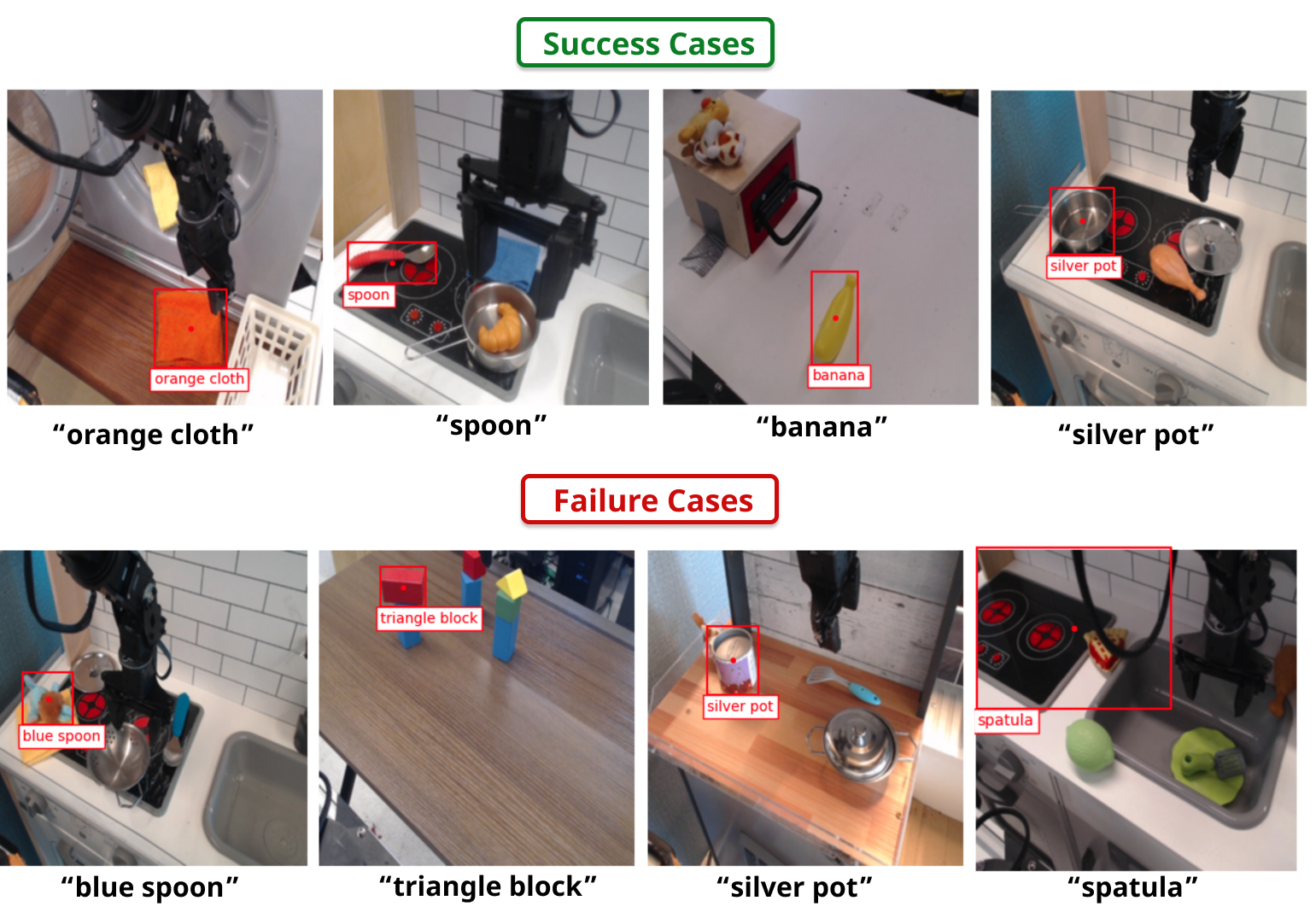}
    \caption{\figtitle{Examples of extracted bounding boxes for MOO}. Despite our best attempts, the bounding boxes extracted at train time are noisy due to the unstructured nature of BridgeData V2. This might have led to MOO's poor performance in our evaluations.}
    \vspace{-0.4cm}
    \label{fig:moo_mask}
\end{figure}

\subsubsection{UniPi}
UniPi~\citep{unipi} trains a video diffusion model $p_\theta \left(\tau \mid \bs_0, l \right)$ to generate a sequence of frames given a language command and an initial frame. The original paper employs the model architecture from Imagen Video~\citep{imagen-video, vdm}.
To achieve higher resolution and longer videos for their real-world results, the authors leverage a 1.7B 3D UNet and four pretrained super-resolution models from Imagen Video, with 1.7B, 1.7B, 1.4B, and 1.2B parameters, respectively. 
Since the original models and code are not publicly available, we tried to replicate their approach in two different ways.

\textbf{UniPi (HiP, ~\citet{hip})} For the second approach, we followed the UniPi replication in ~\citet{hip}. We trained a latent video diffusion model from PVDM~\citep{pvdm}, building upon the codebase \url{https://github.com/sihyun-yu/PVDM} where we added first-frame conditioning. We first trained the video autoencoder to project a video of size $16  \times 128  \times 128$ into a latent space, followed by training a PVDM-L model that uses a 2D UNet architecture. We used a Flan-T5-Base~\citep{flan-t5} encoder to obtain the language embeddings.

\textbf{UniPi (Ours).} We implemented a 3D UNet video diffusion model, following~\citet{vdm, imagen-video} and adding UniPi's first-frame conditioning. Due to limited compute, we did not train spatial/temporal super-resolution models; instead, we trained a 3D UNet-based diffusion model to directly generate images with a resolution of \(128 \times 128\). The model includes 4 residual blocks, with (input channels, output channels) as follows: $(64, 64)$, $(64, 128)$, $(128, 256)$, and $(256, 640)$. The model is trained to produce the trajectory with a fixed horizon of 10 frames $\tau_t = \left\{ s_t, s_{t+1}, \ldots, s_{t+9} \right\}$ conditioned on the current frame $s_t$ and language command. We used a frozen pretrained CLIP~\citep{clip} encoder to obtain the language embeddings.

\textbf{Data and training details.} Following the UniPi baseline from HiP~\citep{hip}, we utilized Ego4D~\citep{grauman2022ego4d} to instill the model with Internet-scale knowledge. For UniPi (Ours), we first pretrained the video model on Ego4D for 270K steps, and finetuned it on the robotics dataset (CALVIN for simulation and BridgeData V2 for the real world) for an additional 200K steps. We used a batch size of 4 during training. For UniPi (HiP), we jointly trained a single model on Ego4D, BridgeData V2, and CALVIN at the same time. The autoencoder was trained for 85K steps, and the PVDM-L model was trained for 200K steps. We used a batch size of 8 during training.

\textbf{Inverse model and test time control.} To extract actions from generated videos, we trained a Gaussian inverse dynamics model $\pi_\phi \left(\ba \mid \bs_{t}, \bs_{t+1} \right)$ to predict the action from two adjacent frames. The model consists of a ResNet-50 image encoder and 3 256-unit MLP layers. During test time, given the current observation $s_t$ and the language command $l$, we synthesize $H$ image frames from the video model and apply the inverse dynamics model to obtain the corresponding $H-1$ actions. The predicted actions are executed, and we generate a new video from $s_{t+H-1}$ and repeat the process.

\textbf{Generated videos.}  While the quality of the video model trained on the simulation dataset is good enough for solving the tasks on the CALVIN benchmark as shown in Table~\ref{tab:calvin}, we found that it is nontrivial to obtain high-quality generations for the real-world scenes. We show examples of generations in Appendix~\ref{appendix:unipi_examples}. 

\subsubsection{Language-conditioned behavior cloning (LCBC)}
\label{appendix:lcbc}
For the LCBC baseline, we use the same architecture and hyperparameters as the low-level policy in \methodname. We encode the language instruction using MUSE~\citep{yangMultilingualUniversalSentence2019} and feed it into the ResNet-50 image encoder using FiLM conditioning~\citep{perezFiLMVisualReasoning2017}. We found that this policy outperformed the LCBC policy used in \citet{bridgev2} and \citet{grif} (which was not a diffusion policy), as well as the RT-1 architecture~\citep{brohan2022rt} trained on BridgeData V2 (also appearing in \citet{bridgev2}).

\section{Experimental Setup}
\label{appendix:experimental_setup}

\textbf{CALVIN.} We follow the evaluation setup from \citet{calvin}. During each trial, the policy gets 360 timesteps total to complete a chain of 5 instructions, only moving on to the next instruction once the previous one is completed. Subsequent instructions in each language chain are chosen by the simulator based on environment affordances after completing the previous language instruction. Results in Table~\ref{tab:calvin} are averaged over 100 trajectories, enabling direct comparison with results from~\citet{ge2023policy}, which are also averaged over 100 trajectories. Results for UniPi (HiP) and UniPi (Ours) were averaged over 25 trajectories, due to the computational cost associated with querying the video diffusion models, and the relatively short video generation horizon (10 timesteps) requiring frequent regeneration.

To obtain comparisons with an oracle GCBC policy in Table~\ref{tab:compare_to_oracle}, we sampled ground truth goal image states from the simulator to feed to the oracle policy by manually resetting the environment internal state to the state corresponding to the goal being achieved. This manual resetting process was only amenable for certain tasks in CALVIN, namely the ones in Table~\ref{tab:compare_to_oracle}. Results in the table correspond to a total of 200 evaluation trajectories. 

\textbf{Real world.} Figure~\ref{fig:scenes} provides an exhaustive list of all 9 tasks we evaluate in the real world. We conduct 10 trials per task and report the average success rates. We vary the positions and orientations of various objects between trials, but use the same set of positions and orientations for evaluating all methods. In Scene A, we change the positions and orientations of the plate, eggplant, and carrot every trial; the pot always remains in the drying rack. For ``put carrot on plate'', we start the carrot in the pot for 4 out of 10 trials, and in the sink otherwise. The eggplant always starts in the sink. In Scene B, we change the positions of the bell pepper, orange pot, and ceramic bowl every trial, performing 5 trials where the target container is closer to the target object and 5 trials where it is further away. In Scene C, for ``put X in the wooden bowl,'' we change the positions of the 3 objects every trial while the wooden bowl remains on the left; we also randomize the rotations of the objects up to 45 degrees from vertical. For ``move the wooden bowl to the top of the table,'' we change the horizontal position of the bowl every trial, but it always starts at the bottom of the table.

\section{Qualtitative Examples of Video Generation}
\begin{figure}[h]
    \centering
    \includegraphics[width=0.8\linewidth]{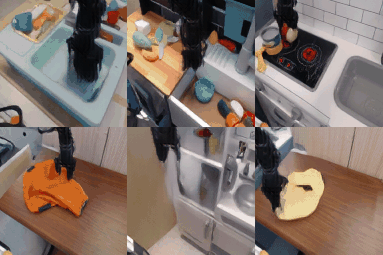}
    \caption{\figtitle{UniPi (HiP) example generations}. Observe that the model is not able to generate a clear robot gripper, and produces videos that are blurry and visually poor.}
    \vspace{-0.4cm}
    \label{fig:pvdm-bridge}
\end{figure}

\begin{figure}[h]
    \centering
    \vspace{-0.2cm}
    \includegraphics[width=\linewidth]{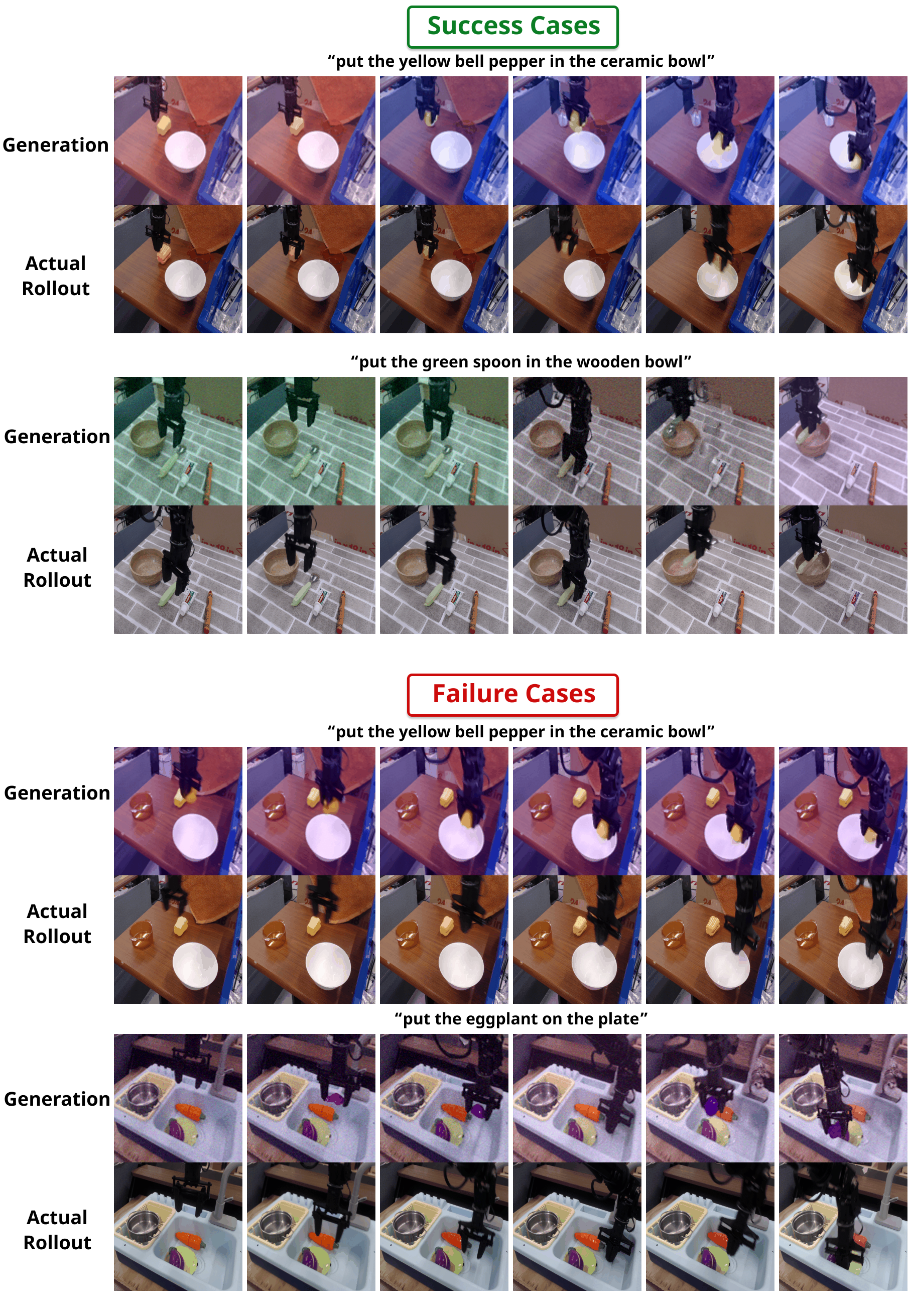}
    \vspace{-0.4cm}
    \caption{\figtitle{UniPi (Ours) example generations and the resulting robot behaviors}. While the robot succeeded in a few trials, we observed that in most cases the generated videos suffered from hallucination and physical inconsistency, which confused the low-level policy.}
    \vspace{-0.4cm}
    \label{fig:unipi-bridge}
\end{figure}

Figure~\ref{fig:pvdm-bridge} shows the video frames generated by the UniPi (HiP) model. The model fails to generate the robot's gripper accurately and consistently produces blurry results. For this reason, we did not evaluate this model on the real robot. Instead, we evaluated UniPi (Ours) model for the real-world tasks. Figure~\ref{fig:unipi-bridge} shows examples of success and failure cases of the test trials. While the robot succeeded in a few trials, we observed that in most cases the generated videos suffered from hallucination and physical inconsistency, which led to UniPi's poor performance.
\label{appendix:unipi_examples}

\clearpage

\section{Qualitative Examples of SuSIE on CALVIN}
\label{appendix:calvin}

\begin{figure}[h]
    \centering
    \vspace{-0.2cm}
    \includegraphics[height=20cm]{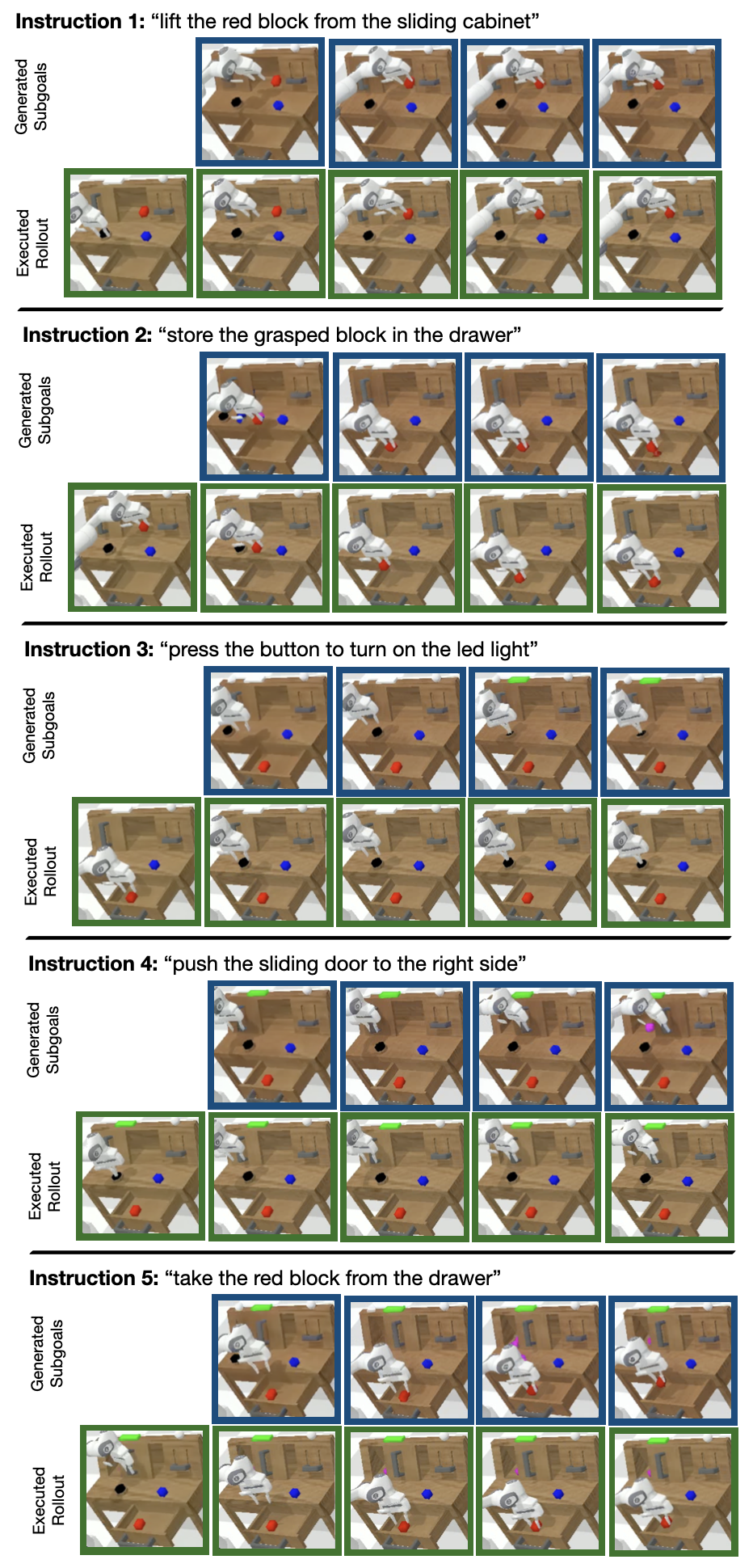}
    \vspace{-0.4cm}
    \caption{\figtitle{Example rollouts in CALVIN.} Visualization of SuSIE-generated subgoals and trajectory rollouts on a successful 5-instruction language chain in CALVIN.}
    \vspace{-0.4cm}
    \label{fig:calvin-language-chain}
\end{figure}

\end{document}